\documentclass[12pt,a4paper]{report}

\usepackage[utf8]{inputenc}
\usepackage[T1]{fontenc}
\usepackage{geometry}
\usepackage{hyperref}
\usepackage{titlesec}
\usepackage{booktabs}
\usepackage{multirow}
\usepackage{xcolor}
\usepackage{amsmath}
\usepackage{amssymb}
\usepackage{graphicx}
\usepackage{dirtree}
\usepackage{float}
\usepackage{makecell}
\usepackage[round]{natbib}

\begin{document}

\pagenumbering{roman}

\begin{titlepage}
    \centering
    \vspace*{1cm}
    \Huge \textbf{Capstone Project: Exploring Convolutional Neural Processes for Weather Downscaling} \\
    \vspace{1.5cm}
    \Large \textbf{Student Name:} Francisco Passos \\
    \Large \textbf{Student ID/LEGI:} 24-952-418 \\
    \vspace{1cm}
    \large \textbf{Course:} DAS in Data Science \\
    \large \textbf{Institution:} ETH Zürich \\
    \vspace{2cm}
    \large \textbf{Submission Date:} February 13, 2026
\end{titlepage}

\section*{Abstract}

Global reanalysis products such as ERA5-Land provide spatially complete weather fields but at resolutions too coarse for local applications, particularly in mountainous regions where temperature can vary by several degrees over short distances. This project investigates Convolutional Conditional Neural Processes (ConvCNPs) for statistical downscaling of daily maximum temperature from the ${\sim}$11\,km ERA5-Land grid to ${\sim}$1\,km resolution over Switzerland, building upon the architecture of \citet{vaughan2022} and adapting it to the topographically complex Swiss domain with high-resolution elevation features from the swisstopo DHM25. The best model, trained on ten years of data (2014--2023) with five-fold temporal cross-validation, achieves a mean absolute error of 1.31\,\textdegree C and a CRPS-based skill score of 0.524 relative to bilinear interpolation, reducing the expected prediction error by more than half. An ablation study reveals that the elevation MLP is the indispensable component---without it, the model diverges entirely---while explicit seasonal features and Topographic Position Index provide secondary benefits. Under sparse on-grid input the model degrades gracefully, maintaining positive skill down to approximately 10\% of the input grid; however, zero-shot deployment on off-grid station observations does not achieve positive skill at any density tested. All configurations exhibit severely overconfident uncertainty estimates, a structural limitation of the Gaussian likelihood training objective. These results demonstrate that ConvCNPs are a viable and effective approach to climate downscaling in complex terrain, and identify uncertainty calibration and native support for non-gridded input as the key challenges for operational deployment.

\section*{Acknowledgements}

\par{
I wish to express my sincere gratitude to my supervisor Dr. Christian Donner, for his invaluable expert guidance and kind support throughout the duration of this project, and to the Swiss Data Science Center, for the opportunity, support and resources to conduct this project. I am also grateful to Ann-Kristin Grimm, Vit Hornacek, Dmitry Fedoruk and Brad Kratochvil, for the support and flexibility to balance my professional responsibilities with my studies. Finally, my deepest thanks go to my wife and son, Isabel and André, for the motivation to pursue this adventure, for their continuous encouragement, and for their patience throughout the demanding periods of my Diploma in Advanced Studies.
}

\tableofcontents
\listoffigures
\listoftables

\section*{List of Acronyms}
\begin{description}
    \item[ANP] Attentive Neural Process
    \item[CNN] Convolutional Neural Network
    \item[CNP] Conditional Neural Process
    \item[ConvCNP/CCNP] Convolutional Conditional Neural Process
    \item[ConvNP] Convolutional Neural Process
    \item[CRPS] Continuous Ranked Probability Score (Probabilistic skill score)
    \item[DEM] Digital Elevation Model
    \item[DHM25] Digital Height Model 25\,m (swisstopo)
    \item[ECMWF] European Centre for Medium-Range Weather Forecasts
    \item[ERA5-Land] ECMWF Reanalysis v5 Land (High-resolution terrestrial dataset)
    \item[GP] Gaussian Process
    \item[MAE] Mean Absolute Error
    \item[MLP] Multi-Layer Perceptron
    \item[NP] Neural Process
    \item[NWP] Numerical Weather Prediction
    \item[RMSE] Root Mean Square Error
    \item[TPI] Topographic Position Index
\end{description}

\newpage
\pagenumbering{arabic}

\chapter{Introduction}
\section{Project Motivation}

Global numerical weather prediction (NWP) models and reanalysis products such as ERA5-Land provide spatially complete representations of atmospheric variables on regular grids. However, their spatial resolution---typically on the order of 9\,km for ERA5-Land---is too coarse for many local applications. Agriculture, infrastructure planning, energy systems, and natural hazard management all require temperature estimates at specific locations, where the local topography, land cover, and altitude can cause conditions to differ substantially from the surrounding grid cell average. In mountainous regions such as Switzerland, where elevation can change by hundreds of meters over short horizontal distances, this mismatch is particularly pronounced.

The traditional approach to bridging this resolution gap is known as \textit{statistical downscaling}: learning a mapping from coarse-resolution gridded fields to fine-resolution local observations. Classical methods range from simple interpolation schemes (e.g.\ bilinear or nearest-neighbor) to more sophisticated geostatistical techniques such as kriging. While effective in some settings, these approaches often struggle to capture the complex, nonlinear relationships between large-scale atmospheric patterns and local surface conditions, particularly when auxiliary information such as high-resolution topography must be incorporated.

Recent advances in deep learning have opened new avenues for statistical downscaling. Among the most promising are Neural Processes \citep[NPs;][]{garnelo2018np}, a family of models that combine the flexibility of neural networks with principled uncertainty quantification. In particular, Convolutional Conditional Neural Processes \citep[ConvCNPs;][]{gordon2019} are designed to operate on spatial data: they accept an arbitrary set of context observations on a grid, learn translation-equivariant representations through convolutional architectures, and produce predictions---with calibrated uncertainty estimates---at any requested target location. \citet{vaughan2022} demonstrated the viability of ConvCNPs for climate downscaling over the United Kingdom, showing competitive performance against established baselines.

This project, defined by Dr.\ Christian Donner at ETH Z\"urich, sets out to explore the applicability of ConvCNPs for downscaling 2-meter temperature over Switzerland. By building upon the codebase and methodology of \citet{vaughan2022}, the goal is to assess how well these models perform in a new geographic and topographic context, to understand the contribution of individual input features, and to evaluate the robustness of the approach when context data is sparse---a scenario that is common in operational settings where station networks are thin or data is partially missing.

\section{Problem Definition and Goal}

The core data science problem addressed in this project is \textit{spatial interpolation with uncertainty}: given a set of coarse-resolution gridded weather fields (context points) and auxiliary topographic information, predict the 2-meter air temperature at specific ground-truth station locations (target points), together with a well-calibrated measure of predictive uncertainty.

Formally, let
\[
    \mathbf{X}_C = \{(\mathbf{x}_i,\; y_i)\}_{i=1}^{N_C}
\]
denote the \textit{context set}, where $\mathbf{x}_i$ are spatial coordinates and $y_i$ are observed values from the ERA5-Land reanalysis grid and associated features (elevation, Topographic Position Index, seasonal encodings). Let
\[
    \mathbf{X}_T = \{\mathbf{x}_j^*\}_{j=1}^{N_T}
\]
denote the \textit{target set}, corresponding to MeteoSwiss ground-truth weather stations. The model learns a mapping $f_\theta$ that, conditioned on $\mathbf{X}_C$, predicts the distribution
\[
    p(y_j^* \mid \mathbf{x}_j^*,\; \mathbf{X}_C)
\]
at each target location. Crucially, this is a \textit{meta-learning} formulation: rather than fitting a single regression function, the model is trained across many tasks---each task being a different day's weather pattern with its own context and target sets---so that at inference time it can rapidly condition on a novel context set and generalize to unseen configurations without retraining. The ConvCNP achieves this in three stages:
\begin{enumerate}
    \item \textbf{Encode:} the context set is mapped onto a discretized grid via a set convolution;
    \item \textbf{Process:} the gridded representation is refined through a convolutional neural network (CNN);
    \item \textbf{Decode:} at each target location, a multi-layer perceptron (MLP) maps the learned representation---together with location-specific features such as station elevation---to the \textit{parameters} of a predictive distribution (e.g.\ the mean $\mu_j$ and variance $\sigma_j^2$ of a Gaussian), thereby providing both a point estimate and a calibrated measure of uncertainty.
\end{enumerate}

This project investigates the following research questions:

\begin{enumerate}
    \item \textbf{Model viability:} Can a ConvCNP, originally demonstrated for the UK in \citet{vaughan2022}, produce accurate and well-calibrated temperature predictions for the topographically complex terrain of Switzerland?
    \item \textbf{Feature importance:} What is the relative contribution of auxiliary input features---specifically seasonal encodings, elevation, and Topographic Position Index---to prediction accuracy? An ablation study systematically removes these features to quantify their impact.
    \item \textbf{Robustness to sparse input:} How does model performance degrade as the number of available context points is progressively reduced? This question directly addresses the practical scenario of deploying such a model in regions with limited observational coverage.
    \item \textbf{Skill over interpolation:} Does the ConvCNP provide measurable improvement (skill) over direct bilinear interpolation of the ERA5-Land grid to station locations?
\end{enumerate}

\section{Reader guide}

The remainder of this report is structured as follows. Chapter~2 provides the theoretical background on Neural Processes and related work. Chapter~3 details the data sources, preprocessing pipeline, and model implementation. Chapter~4 presents experimental results including global performance metrics, ablation analyses, and sparse-input experiments. Chapter~5 concludes with a discussion of limitations and directions for future work.

For readers primarily interested in practical outcomes and reproducibility, the following entry points may be useful:

\begin{itemize}
    \item \textbf{Headline accuracy and skill:} Table~\ref{tab:full_grid_overview} (full-grid metrics) and Table~\ref{tab:ablation_overview} (ablation results) in Chapter~4.
    \item \textbf{Sparse-input robustness:} Tables~\ref{tab:sparse_era5} and~\ref{tab:sparse_meteoswiss} compare on-grid and off-grid performance at varying context densities (Section~\ref{sec:sparse}).
    \item \textbf{Repository layout and code structure:} Appendix~\ref{sec:code-structure}.
    \item \textbf{How to train and evaluate a model:} Appendix~\ref{sec:training-predicting}, which covers configuration, running the notebooks, and interpreting outputs.
    \item \textbf{Extending the model with new features:} Appendix~\ref{sec:adding-features} documents the three extension points (CNN channels, per-location MLP features, per-timestep MLP features) with code examples.
\end{itemize}
\chapter{Background}
\section{Theoretical Foundations}

This section introduces the Neural Process family of models, building from the foundational ideas to the specific architecture used in this project.

\subsection{From Gaussian Processes to Neural Processes}

Gaussian Processes (GPs) are a classical tool for spatial interpolation. A GP defines a distribution over functions: given a set of observed input--output pairs, it produces a predictive distribution at new locations that is analytically tractable and naturally uncertainty-aware. However, GPs scale poorly with the number of observations---exact inference is $\mathcal{O}(N^3)$ in the number of context points---and encoding complex, nonlinear relationships requires careful kernel engineering.

Neural Processes (NPs), introduced by \citet{garnelo2018np}, were designed to combine the strengths of GPs and neural networks. Like a GP, an NP defines a conditional distribution over functions given a context set. Like a neural network, it learns flexible representations from data and scales efficiently at inference time. The key idea is meta-learning: the model is trained on many tasks (each a different context--target split drawn from a data-generating process), so that it learns to map any context set to a predictive distribution over target values in a single forward pass, without the costly matrix inversions of GPs.

\subsection{Conditional Neural Processes}

The Conditional Neural Process (CNP), also introduced by \citet{garnelo2018cnp}, is the simplest member of the NP family. Given a context set $\{(\mathbf{x}_i, y_i)\}_{i=1}^{N_C}$, a CNP:
\begin{enumerate}
    \item \textbf{Encodes} each context pair independently through a shared encoder network $h_\theta$, producing a representation $\mathbf{r}_i = h_\theta(\mathbf{x}_i, y_i)$;
    \item \textbf{Aggregates} the individual representations into a single, fixed-size summary $\mathbf{r} = \frac{1}{N_C}\sum_{i=1}^{N_C} \mathbf{r}_i$ via a permutation-invariant operation (mean pooling);
    \item \textbf{Decodes} at each target location $\mathbf{x}^*$ by passing $[\mathbf{r},\, \mathbf{x}^*]$ through a decoder network that outputs the parameters of a predictive distribution.
\end{enumerate}
The CNP is trained by maximizing the log-likelihood of the target observations under the predicted distributions, across many tasks. Because each task presents a different context set, the model learns to condition on arbitrary observation configurations---the hallmark of meta-learning.

A limitation of the CNP is that the global aggregation step compresses all spatial information into a single vector $\mathbf{r}$, which can be a bottleneck when the underlying function has rich spatial structure. The Attentive Neural Process (ANP), proposed by \citet{kim2019}, addresses this by using attention mechanisms to produce target-specific representations, but at increased computational cost.

\subsection{Convolutional Conditional Neural Processes}

The Convolutional Conditional Neural Process (ConvCNP), introduced by \citet{gordon2019}, takes a different approach to preserving spatial structure. Instead of aggregating context into a global vector, the ConvCNP places the context observations onto a discretized grid using a \textit{set convolution}---a continuous convolution that smoothly maps irregularly spaced observations to a regular lattice. This gridded representation is then processed by a standard convolutional neural network (CNN), which is naturally suited to extracting spatial features at multiple scales.

The architecture consists of three stages:
\begin{enumerate}
    \item \textbf{Set convolution (Encoder):} Each context observation $(\mathbf{x}_i, y_i)$ contributes to the grid through a kernel centered at $\mathbf{x}_i$. A parallel \textit{density channel} tracks the observation density at each grid cell, providing the model with information about where context data is available and where it is absent. The result is a multi-channel gridded input to the CNN.

    \item \textbf{CNN (Processor):} A deep convolutional network---typically a ResNet with skip connections---processes the gridded representation. Because convolutions are translation-equivariant, the CNN treats all spatial locations consistently: a weather pattern produces the same learned features regardless of where it occurs in the domain.

    \item \textbf{Decoder:} At each target location, the CNN output is interpolated and passed through a decoder that outputs the parameters of the predictive distribution. In the variant used in this project, the decoder is a multi-layer perceptron (MLP) that also accepts location-specific auxiliary features (e.g.\ elevation), enabling the model to correct for local effects that the coarse CNN representation cannot capture.
\end{enumerate}

The ConvCNP's translation equivariance is a key inductive bias for spatial problems: it encodes the prior knowledge that physical laws do not depend on absolute position. Combined with the set convolution's ability to handle arbitrary and variable-size context sets, this makes the ConvCNP well suited to weather downscaling, where the context may be a regular reanalysis grid, a sparse station network, or anything in between.

\section{Related Work}

\subsection[Vaughan et al. (2022): ConvCNPs for Climate Downscaling]{\citet{vaughan2022}: ConvCNPs for Climate Downscaling}

\citet{vaughan2022} were the first to apply ConvCNPs to the problem of climate downscaling, demonstrating the approach for temperature and precipitation over the United Kingdom. Their work used ERA5 reanalysis data (at approximately 27\,km resolution) as context, with ground-truth observations from the VALUE station network as targets. The model included multiple ERA5 channels---temperature, pressure, humidity, wind components, and geopotential---providing a rich atmospheric context.

Their results showed that ConvCNPs could match or exceed the performance of established statistical and dynamical downscaling baselines, while providing well-calibrated uncertainty estimates and requiring no hand-crafted feature engineering beyond the choice of input channels. The accompanying open-source codebase provided implementations of the encoder, CNN, MLP decoder, training loop, and evaluation metrics.

\subsection[How this project extends Vaughan et al. (2022)]{How this project extends \citet{vaughan2022}}

This project builds directly on the \citet{vaughan2022} codebase and methodology, adapting and extending it for a new geographic domain and research focus. The key differences are summarized below.

\paragraph{Geographic domain.} The study area shifts from the relatively flat UK mainland (elevation range 0--1000\,m) to Alpine Switzerland (elevation range approximately 180--4675\,m). The extreme topographic complexity of the Swiss Alps---where elevation can change by over 1000\,m within a few kilometers---presents a substantially harder downscaling challenge and makes elevation-aware modeling essential.

\paragraph{Input data: ERA5-Land.} Where \citet{vaughan2022} used ERA5 at $\sim$27\,km resolution, this project uses ERA5-Land at 0.1\textdegree{} ($\sim$11\,km) resolution. ERA5-Land is the ECMWF's high-resolution land-surface reanalysis, providing finer spatial detail over terrestrial areas. To keep the exploration focused and due to the limited duration of the capstone project, a reduced set of input channels is used: normalized 2-meter maximum temperature, latitude, longitude, and cyclical day-of-year encodings (cosine and sine), for a total of five CNN input channels.

\paragraph{Target data: MeteoSwiss observational grid.} Rather than predicting at scattered station locations, this project targets the MeteoSwiss gridded observational product (TmaxD v2.0), which provides daily maximum temperature on a regular 240$\times$370 grid in the Swiss LV95 coordinate system. This dense target grid enables spatially comprehensive evaluation but requires coordinate transformations between the WGS84 system of ERA5-Land and the Swiss LV95 projection.

\paragraph{Elevation handling and the missing geopotential channel.} \citet{vaughan2022} included geopotential---directly related to altitude---as one of the CNN input channels, meaning their convolutional backbone was not elevation-agnostic (despite the paper mentioning it is, likely a description that stayed from an earlier version of the model). This was not initially recognized during the development of the present project, and as a result geopotential was \textit{not} included as a CNN channel. Had this been noticed earlier, it would have been included to achieve parity. This omission is acknowledged as a regression: the CNN in this project lacks direct elevation awareness at the grid level, placing a greater burden on the downstream MLP to compensate.

That said, the project does provide elevation information to the model through a dedicated MLP pathway. The MLP receives the CNN's base prediction together with three high-resolution topographic features computed at each target location:
\begin{itemize}
    \item \textbf{True elevation}: interpolated from a 25\,m-resolution digital elevation model (DEM);
    \item \textbf{Elevation difference}: the difference between the true elevation and the ERA5-Land grid cell elevation, capturing the local altitude mismatch that the coarse grid cannot resolve;
    \item \textbf{Topographic Position Index (TPI)}: a measure of relative elevation compared to the surrounding terrain at a 500\,m scale, indicating whether a location sits on a ridge, in a valley, or on a slope.
\end{itemize}
While this MLP-based correction allows the model to account for local topographic effects at high resolution, it cannot fully replace the spatial elevation context that a CNN channel would provide. Including geopotential as a CNN channel remains a clear avenue for improvement (see Chapter~5).

\paragraph{Explicit seasonal encoding.} In addition to the cyclical day-of-year features in the CNN input channels, the same seasonal features are optionally passed directly to the elevation MLP. This enables the MLP to learn season-dependent lapse rates---for instance, steeper temperature gradients with altitude in summer versus temperature inversions in winter valleys---without relying solely on the CNN to propagate this information.

\paragraph{Focus on operationalization.} Beyond model accuracy, this project investigates practical deployment considerations: how the model performs with progressively sparser context data, the relative importance of individual input features (via ablation studies), and the skill gained over simple ERA5-Land interpolation. These questions are directly relevant to assessing whether ConvCNPs can be operationalized for routine downscaling in data-sparse regions.
\chapter{Methodology}
\section{Data Acquisition and Description}

Three primary data sources are used in this project, summarized in Table~\ref{tab:datasets}.

\begin{table}[ht]
\centering
\caption{Summary of datasets used in this project.}
\label{tab:datasets}
\begin{tabular}{@{}lll@{}}
\toprule
\textbf{Dataset} & \textbf{Role} & \textbf{Format} \\
\midrule
ERA5-Land         & Context input      & NetCDF \\
MeteoSwiss TmaxD  & Ground truth       & NetCDF \\
swisstopo DHM25   & Elevation features & Zarr \\
\bottomrule
\end{tabular}
\end{table}

\subsection{ERA5-Land (Context)}

ERA5-Land \citep{era5land} is the high-resolution land-surface component of the ECMWF ERA5 reanalysis, providing hourly fields on a regular 0.1\textdegree{} latitude--longitude grid from 1950 to the present. Data is downloaded from the Copernicus Climate Data Store as yearly NetCDF files.

\begin{itemize}
    \item \textbf{Variables:} daily maximum 2-meter air temperature (\texttt{t2m\_max}) and the static geopotential field (\texttt{z}).
    \item \textbf{Grid:} 29 latitude $\times$ 61 longitude points over Switzerland (1\,769 cells, $\sim$11\,km spacing).
    \item \textbf{Temporal ranges:} a single year (2023, 365 days) for rapid iteration, and a ten-year window (2014--2023, $\sim$3\,650 days) for the primary experiments.
\end{itemize}

The geopotential field is converted to altitude via $h = z / g$, where $g = 9.80665$\,m/s\textsuperscript{2}. This provides the coarse grid-cell elevation used to compute the elevation difference feature (see Section~\ref{sec:feature_engineering}).

\subsection{MeteoSwiss TmaxD v2.0 (Ground Truth)}

The target observations come from the MeteoSwiss gridded temperature product TmaxD v2.0 \citep{meteoswiss_tmax}, derived from station observations using spatial interpolation.

\begin{itemize}
    \item \textbf{Variable:} daily maximum temperature.
    \item \textbf{Grid:} 240$\times$370 points in the Swiss LV95 coordinate system (EPSG:2056), $\sim$1\,km spacing.
    \item \textbf{Units:} degrees Celsius, converted to Kelvin to match the ERA5-Land reference frame before normalization.
    \item \textbf{Coverage:} after flattening, 88\,800 points total, of which a subset are valid (non-NaN) on any given day---points outside the Swiss border are masked.
\end{itemize}

\subsection{swisstopo DHM25 high-Resolution Topography}

Elevation and terrain features are derived from the swisstopo DHM25 \citep{dhm25} (Digital Height Model), stored locally in Zarr format on a 25\,m Swiss LV95 grid (12\,800$\times$14\,200 pixels). The dataset includes pre-computed terrain derivatives.

\begin{itemize}
    \item \textbf{DEM}: surface elevation in meters.
    \item \textbf{TPI\_500M}: Topographic Position Index at a 500\,m radius, quantifying relative elevation compared to the surrounding terrain.
\end{itemize}

These fields are interpolated to target point locations using bilinear interpolation, after converting target coordinates from WGS84 to the LV95 projection.

\section{Data Preprocessing and Exploration}
\label{sec:feature_engineering}

\begin{figure}[ht]
    \centering
    \includegraphics[width=0.85\textwidth]{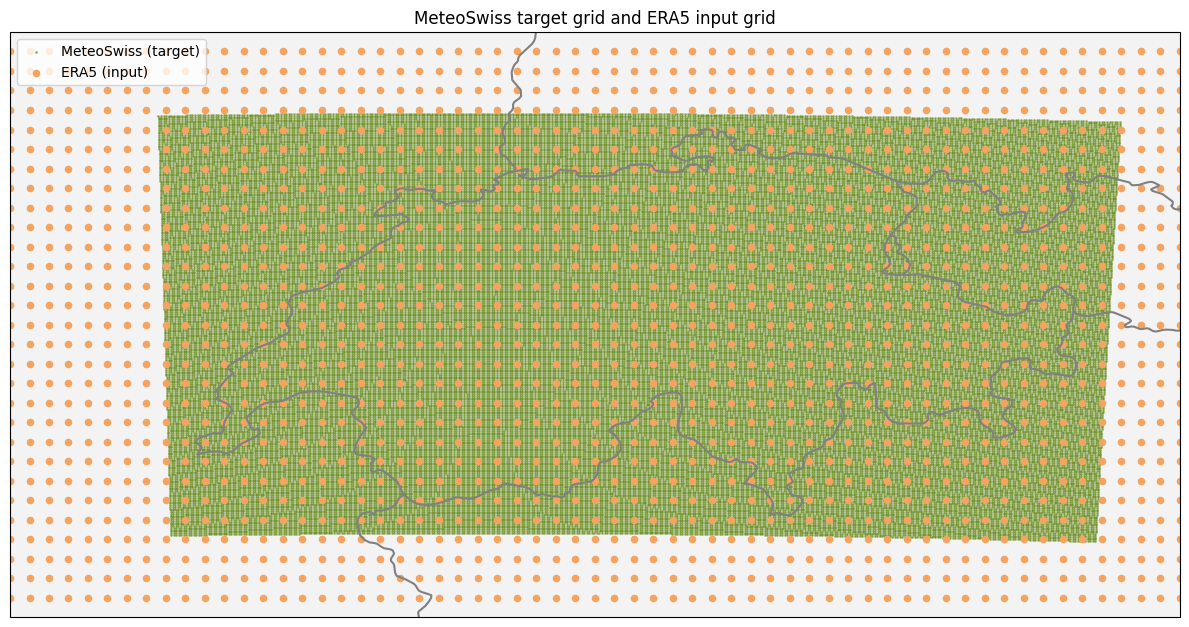}
    \caption{Overlay of the ERA5-Land context grid (orange, 29$\times$61 points at $\sim$11\,km spacing) and the MeteoSwiss target grid (green, 240$\times$370 points at $\sim$1\,km spacing). The resolution gap---roughly a factor of 10 in each spatial dimension---is the core downscaling challenge.}
    \label{fig:grids}
\end{figure}

The preprocessing pipeline transforms the three raw datasets into the tensor representations consumed by the ConvCNP. Because the context input (ERA5-Land) and the target observations (MeteoSwiss) live on different grids, in different coordinate systems, and in different units, the main challenges are coordinate alignment, consistent normalization, and the construction of auxiliary features that help the model account for topographic effects. Figure~\ref{fig:grids} illustrates the spatial relationship between the two grids.

\subsection{Data Quality and Cleaning}

ERA5-Land is a reanalysis product: it is generated by assimilating historical observations into a numerical weather model under rigorous quality control by ECMWF. As a result, the data is spatially and temporally complete---there are no missing values, no gaps in the time series, and no outliers arising from instrument failures. Similarly, the MeteoSwiss TmaxD product is a quality-controlled gridded interpolation of station observations, and the swisstopo DHM25 is a professionally produced digital elevation model. Consequently, no ad-hoc cleaning, gap-filling, or outlier removal was required. The only data filtering applied was temporal subsetting: selecting either a single year (2023, 365 days) for rapid prototyping or a ten-year window (2014--2023, $\sim$3\,650 days) for the primary experiments.

\subsection{Coordinate Alignment and Regridding}

The three datasets use two different coordinate reference systems. ERA5-Land uses a regular latitude--longitude grid in WGS84 (EPSG:4326), while both the MeteoSwiss observations and the swisstopo DEM use the Swiss LV95 projected coordinate system (EPSG:2056), whose axes are expressed in meters. Aligning these required two types of coordinate transformations, both performed using \texttt{pyproj}:

\begin{enumerate}
    \item \textbf{MeteoSwiss $\rightarrow$ ERA5 frame.} The MeteoSwiss dataset provides latitude and longitude fields in WGS84 degrees alongside its native LV95 grid indices. These WGS84 coordinates are normalized to $[0, 1]$ using the ERA5-Land bounding box, so that target locations are expressed in the same reference frame as the context grid.

    \item \textbf{WGS84 $\rightarrow$ LV95 for DEM lookup.} To extract high-resolution elevation at each MeteoSwiss target point, the WGS84 target coordinates are transformed to LV95 meters, then used to query the swisstopo DEM and TPI fields via bilinear interpolation (\texttt{xarray.DataArray.interp}).
\end{enumerate}

The ERA5-Land grid-cell elevation (derived from the geopotential field, see below) is similarly interpolated to target locations, but this interpolation uses the native WGS84 coordinates of the ERA5 grid since both the geopotential and the target latitude/longitude are in the same coordinate system.

\subsection{Feature Engineering}
\label{sec:feature_engineering_details}

\subsubsection{Temperature Normalization}

Both the ERA5-Land context temperatures and the MeteoSwiss target temperatures are normalized using a global Z-score computed from the ERA5-Land training data:
\begin{equation}
    \hat{T} = \frac{T - \mu_{\text{ERA5}}}{\sigma_{\text{ERA5}}}
\end{equation}
where $\mu_{\text{ERA5}}$ and $\sigma_{\text{ERA5}}$ are the mean and standard deviation of the ERA5-Land \texttt{t2m\_max} field over the entire training period. Using the \textit{same} normalization for both context and target ensures that the model operates in a consistent scale. The MeteoSwiss data, originally in degrees Celsius, is first converted to Kelvin before applying the normalization.

\subsubsection{Coordinate Normalization}

Latitude and longitude are min--max normalized to $[0, 1]$ using the ERA5-Land bounding box:
\begin{equation}
    \hat{\phi} = \frac{\phi - \phi_{\min}}{\phi_{\max} - \phi_{\min}}, \qquad
    \hat{\lambda} = \frac{\lambda - \lambda_{\min}}{\lambda_{\max} - \lambda_{\min}}
\end{equation}
These normalized coordinates serve two roles: they form two of the five CNN input channels (enabling the model to learn position-dependent patterns), and they define the target locations passed to the decoder.

\subsubsection{Cyclical Seasonal Encoding}

The day of year $d \in \{1, \ldots, 365\}$ is encoded as a pair of cyclical features:
\begin{equation}
    c_{\cos} = \cos\!\left(\frac{2\pi(d-1)}{365}\right), \qquad
    c_{\sin} = \sin\!\left(\frac{2\pi(d-1)}{365}\right)
\end{equation}
This representation avoids the discontinuity that a raw day-of-year integer would introduce at the year boundary (day 365 $\rightarrow$ day 1). These two channels are broadcast to the full spatial grid and appended to the ERA5-Land input tensor as channels 4 and 5. Additionally, the same seasonal features are optionally fed directly to the elevation MLP, allowing it to learn season-dependent lapse rate corrections (e.g.\ steeper gradients in summer versus temperature inversions in winter valleys).

\subsubsection{Altitude from Geopotential}

The ERA5-Land geopotential field $\Phi$ (in m$^2$\,s$^{-2}$) is converted to grid-cell altitude via:
\begin{equation}
    h_{\text{grid}} = \frac{\Phi}{g}, \qquad g = 9.80665 \;\text{m\,s}^{-2}
\end{equation}
This provides a coarse ($\sim$11\,km) estimate of the terrain height at each ERA5-Land grid cell. Since the geopotential is static, only the first time step is used. The resulting altitude field is interpolated to target locations via bilinear interpolation and used to compute the elevation difference feature described below.

\subsubsection{High-Resolution Topographic Features}

Three elevation-related features are computed at each MeteoSwiss target point and passed to the elevation MLP:

\begin{enumerate}
    \item \textbf{True elevation} ($h_{\text{true}}$): bilinearly interpolated from the 25\,m swisstopo DEM at the target's LV95 coordinates.

    \item \textbf{Elevation difference} ($\Delta h = h_{\text{true}} - h_{\text{grid}}$): the difference between the true elevation and the ERA5-Land grid-cell elevation interpolated to the same point. This captures the local altitude mismatch that the coarse grid cannot resolve---for instance, a mountain peak that lies within a grid cell whose mean elevation is hundreds of meters lower.

    \item \textbf{Topographic Position Index} (TPI$_{500}$): interpolated from the swisstopo TPI at a 500\,m radius, quantifying whether a target point sits on a ridge (positive TPI), in a valley (negative TPI), or on an open slope (near-zero TPI). This encodes local terrain shape beyond what a single elevation value conveys.
\end{enumerate}

These three features are stacked into a per-point elevation tensor of shape (\textit{n\_points}, 3), which the MLP uses to adjust the CNN's base prediction for local topographic effects.

\subsection{Context Tensor Assembly}

The final ERA5-Land context tensor has shape (\textit{time}, 5, 29, 61), where the five channels are:
\begin{enumerate}
    \item Normalized daily maximum temperature $\hat{T}$
    \item Normalized latitude $\hat{\phi}$
    \item Normalized longitude $\hat{\lambda}$
    \item Cosine seasonal encoding $c_{\cos}$
    \item Sine seasonal encoding $c_{\sin}$
\end{enumerate}
A pre-computed distance tensor of shape (\textit{n\_target\_points}, 29, 61) stores the squared Euclidean distance from each target location to every ERA5-Land grid cell. This tensor is used by the set convolution encoder to weight context observations according to their proximity to each target.

\subsection{Metadata Preservation for Reproducibility}

All normalization statistics---the global mean and standard deviation of the ERA5-Land temperature field, the latitude and longitude bounding box, and the original coordinate arrays---are encapsulated in an \texttt{Era5Metadata} dataclass and serialized to JSON alongside each trained model. Similarly, the full training configuration (hyperparameters, ablation switches, data paths, and random seeds) is captured in a \texttt{Params} dataclass and persisted in the same manner. This ensures that any trained model can be loaded and used for inference with the exact normalization and configuration parameters that were used during training, without relying on the original data files being accessible. Both dataclasses support round-trip JSON serialization, including special handling for NumPy arrays in the metadata.

\section{Model Selection and Implementation}

\subsection{Why a Convolutional Conditional Neural Process?}

The downscaling task can be approached with methods ranging from bilinear interpolation to geostatistical kriging to fully convolutional super-resolution networks. The ConvCNP was selected for four reasons:

\begin{enumerate}
    \item \textbf{Probabilistic output.} The model predicts the parameters of a Gaussian distribution $(\mu, \sigma)$ per target point, providing calibrated uncertainty estimates alongside point predictions.
    \item \textbf{Translation equivariance.} The convolutional architecture encodes the prior that spatial relationships are position-invariant---a sound inductive bias for meteorological fields.
    \item \textbf{Flexible context conditioning.} The set convolution encoder accepts variable numbers of context observations at arbitrary locations, making the model robust to missing data.
    \item \textbf{End-to-end learning.} All components are jointly trained by maximizing the log-likelihood of target observations, learning optimal internal representations directly from data.
\end{enumerate}

\subsection[Leveraging Vaughan et al. (2022)]{Leveraging \citet{vaughan2022}}

This project builds on the open-source ConvCNP implementation of \citet{vaughan2022} \citep{vaughan2022code}, which provided the encoder, ResNet decoder, MLP, training loop, loss functions, and evaluation metrics. Several elements had been adapted from Dubois' Neural Process repository \citep{dubois2020npf}.

Starting from this base, very few modifications were made to the code from \citet{vaughan2022}; a detailed list is included in the Appendix. Those python modules were used to build a separate set of libraries and tooling to adapt the model to the Swiss domain:

\begin{itemize}
    \item WGS84/LV95 coordinate handling;
    \item using high-resolution topographic features;
    \item using cyclical seasonal encodings.
\end{itemize}

\subsection{Architecture Overview}
\label{sec:architecture_overview}

The temperature downscaling model (\texttt{TmaxBiasConvCNPElev}, from \citet{vaughan2022}) consists of four sequential stages:

\begin{enumerate}
    \item \textbf{Encoder (set convolution).} Maps the context tensor and a binary mask onto a representation per grid cell using convolutions (kernel size 5).

    \item \textbf{CNN decoder (ResNet).} Residual blocks with convolutions process the encoded grid, compressing the representation.

    \item \textbf{Grid-to-point interpolation.} An MLP maps the CNN output to $(\mu, \sigma)$ at each grid cell. These are interpolated to the target points via an RBF kernel with learnable length scale $\ell$, using a pre-computed distance tensor. The standard deviation is constrained to be strictly positive.

    \item \textbf{Elevation MLP.} The interpolated $(\mu, \sigma)$ are concatenated with five auxiliary features---true elevation, elevation difference, TPI, and two seasonal encodings---and passed through an MLP that outputs topography- and season-corrected predictions.
\end{enumerate}

\subsection{Loss Function}

The model is trained by maximizing the Gaussian log-likelihood:
\begin{equation}
    \mathcal{L} = \frac{1}{N}\sum_{i=1}^{N} \log \mathcal{N}\bigl(y_i \mid \mu_i,\, \sigma_i\bigr)
\end{equation}
where $y_i$ is the observed normalized temperature and $(\mu_i, \sigma_i)$ are the predicted distribution parameters. NaN targets (points outside the Swiss border) are excluded. If we look at the following expansion:

\begin{equation}
    \log \mathcal{N}(y \mid \mu, \sigma) = -\tfrac{1}{2}\log(2\pi) - \underbrace{\log\sigma}_{\text{penalizes low $\sigma$ (overconfidence)}} - \underbrace{\frac{(y - \mu)^2}{2\sigma^2}}_{\text{penalizes high $\sigma$ (underconfidence)}}
\end{equation}

we realize that this objective incentivizes both accurate means and well-calibrated uncertainty: overconfident predictions are penalized by the $\log\sigma$ term, while underconfident predictions are penalized by the squared-error term.

\subsection{Experimental Setup}
\label{sec:experimental_setup}

\subsubsection{Temporal $k$-Fold Cross-Validation}

The dataset is split temporally: a 5-fold scheme divides the time series into contiguous blocks, holding out one block for validation per fold (${\sim}$73 days each for the single-year dataset). For each fold, training days are shuffled and batched, the held-out fold is evaluated after every epoch (NLL, MAE, Pearson and Spearman correlations), the best checkpoint is saved by validation NLL, and early stopping halts training after 10 epochs without improvement.

\subsubsection{Hyperparameters}

Table~\ref{tab:hyperparams} summarizes the hyperparameters. Values marked $\dagger$ follow \citet{vaughan2022}.

\begin{table}[ht]
\centering
\caption{Model and training hyperparameters.}
\label{tab:hyperparams}
\begin{tabular}{@{}llr@{}}
\toprule
\textbf{Category} & \textbf{Parameter} & \textbf{Value} \\
\midrule
\multirow{4}{*}{Architecture}
    & CNN channels$^\dagger$            & 128 \\
    & ResConvBlocks$^\dagger$           & 6 \\
    & Kernel size$^\dagger$             & 5 \\
    & RBF length scale$^\dagger$        & 0.1 \\
\midrule
\multirow{2}{*}{MLP}
    & Hidden layers$^\dagger$           & 4 \\
    & Units per layer$^\dagger$         & 64 \\
\midrule
\multirow{5}{*}{Training}
    & Learning rate$^\dagger$           & $5 \times 10^{-4}$ \\
    & Optimizer                         & Adam \\
    & Batch size                        & 16 (local) / 8 (Renku) \\
    & Max epochs                        & 10 / 30 (default) / 100 \\
    & Early stopping patience           & 10 \\
\midrule
\multirow{2}{*}{Validation}
    & Folds                             & 5 \\
    & Split strategy                    & Temporal (contiguous) \\
\midrule
Reproducibility & Seed                  & 42 \\
\bottomrule
\end{tabular}
\end{table}

Individual models may have been trained with different hyperparameters.
For each model, all hyperparameters were stored in a file named \newline \texttt{trained\_models/<model\_name>/metadata.json}.

\subsection{Ablation Experiments}

To isolate the contribution of individual components, ablation experiments toggle feature flags in the configuration while keeping all other hyperparameters and cross-validation splits identical:

\begin{itemize}
    \item \textbf{No MTPI} (\texttt{USE\_MTPI=False}): MTPI is excluded from the elevation MLP.
    \item \textbf{No elevation} (\texttt{USE\_ELEVATION=False}): the elevation MLP receives only the CNN's base prediction, isolating the CNN's downscaling skill fully from all topographic correction.
    \item \textbf{No seasonal in MLP} (\texttt{SEASONAL\_FEATURES\_IN\_MLP=False}): seasonal features remain in the CNN input but are withheld from the elevation MLP.
    \item \textbf{No seasonal} (\texttt{SEASONAL\_FEATURES=False}): cyclical day-of-year channels are removed entirely, forcing the model to make its predictions without knowing which time in the year it is. Weather behaviour at altitude varies greatly depending on the seasons.
\end{itemize}

\section{Models Trained and Metrics Captured}
\label{sec:models_trained}

All trained models reside under \texttt{trained\_models/}, with each directory name encoding the key experimental variables following the convention:
\begin{center}
\texttt{final-\textit{YYYY}-\textit{N}e-\textit{K}f\,[-ablation-\textit{component}]}
\end{center}
\noindent where \textit{YYYY} is the first year of training data (all datasets end in 2023), \textit{N} is the maximum number of training epochs, \textit{K} is the number of cross-validation folds, and the optional \texttt{-ablation-\textit{component}} suffix identifies the disabled feature. For example, \texttt{final-2014-30e-5f} denotes a model trained on 2014--2023 data for up to 30~epochs with 5-fold temporal cross-validation and all features enabled.

Nine model configurations were trained in total, organized into two groups: \textit{primary models} that vary data volume and training duration with all features enabled, and \textit{ablation models} that each disable one component to isolate its contribution. All models share the architecture and hyperparameters listed in Table~\ref{tab:hyperparams} (Section~\ref{sec:experimental_setup}), differing only in the parameters described below.

\subsection{Primary Models}

The primary experiments study two axes: the effect of training duration (10, 30, and 100~epochs on a single year of data) and the effect of data volume (1~year versus 10~years at a fixed epoch budget). Table~\ref{tab:primary_models} summarizes the four completed configurations.

\begin{table}[ht]
\centering
\caption{Primary model configurations. All use the full feature set (elevation MLP, TPI, seasonal encoding in both CNN and MLP) and 5-fold temporal cross-validation.}
\label{tab:primary_models}
\begin{tabular}{@{}lrrrr@{}}
\toprule
\textbf{Model name} & \textbf{Data range} & \textbf{Days} & \textbf{Epochs} & \textbf{Batch size} \\
\midrule
\texttt{final-2023-10e-5f}  & 2023 only    & 365    & 10  & 8 \\
\texttt{final-2023-30e-5f}  & 2023 only    & 365    & 30  & 8 \\
\texttt{final-2023-100e-5f} & 2023 only    & 365    & 100 & 8 \\
\texttt{final-2014-30e-5f}  & 2014--2023   & 3\,652 & 30  & 8 \\
\bottomrule
\end{tabular}
\end{table}

A fifth configuration, \texttt{final-2004-30e-5f}, attempted to extend the training window to twenty years (2004--2023). The larger dataset required reducing the batch size from~8 to~4 to fit within GPU memory. Only one of five cross-validation folds completed before the computational budget was exhausted; this model was not carried forward to evaluation and is excluded from subsequent results.

\subsection{Ablation Models}

To isolate the contribution of individual model components, four ablation experiments were conducted on the ten-year dataset (2014--2023, 30~epochs, 5~folds). Each ablation disables exactly one feature relative to the fully-featured baseline (\texttt{final-2014-30e-5f}). Table~\ref{tab:ablation_models} lists the configurations and indicates which features are enabled~(\checkmark) or disabled~(--) in each variant.

\begin{table}[ht]
\centering
\caption{Ablation model configurations. All use 2014--2023 data, 30~epochs, 5~folds, batch size~8. The baseline row repeats the fully-featured ten-year model for reference.}
\label{tab:ablation_models}
\begin{tabular}{@{}lcccc@{}}
\toprule
\textbf{Ablation suffix} & \textbf{Elev.} & \textbf{TPI} & \textbf{Season. (CNN)} & \textbf{Season. (MLP)} \\
\midrule
(none---baseline)                  & \checkmark & \checkmark & \checkmark & \checkmark \\
\texttt{-ablation-elevation}       & --         & --         & \checkmark & \checkmark \\
\texttt{-ablation-mtpi}            & \checkmark & --         & \checkmark & \checkmark \\
\texttt{-ablation-seasonal-all}    & \checkmark & \checkmark & --         & --         \\
\texttt{-ablation-seasonal-in-mlp} & \checkmark & \checkmark & \checkmark & --         \\
\bottomrule
\end{tabular}
\end{table}

Disabling elevation (\texttt{USE\_ELEVATION=False}) also disables TPI, since TPI is an input to the elevation MLP (Section~\ref{sec:architecture_overview}). This variant therefore removes the entire topographic correction stage, and the model's output relies solely on the CNN decoder's grid-to-point interpolation.

\subsection{Metrics Captured}

Each evaluation run---covering all five cross-validation folds---produces a JSON file following the naming convention \texttt{metrics-\textit{INPUT}-\textit{SPARSITY}-\textit{value}.json} (see Appendix for the full specification). The recorded metrics fall into four groups:

\begin{description}
    \item[Point accuracy.] Mean Absolute Error (MAE), Root Mean Squared Error (RMSE), and mean bias, all in~\textdegree C.

    \item[Probabilistic skill.] The Continuous Ranked Probability Score (CRPS) and a skill score relative to deterministic ERA5-Land bilinear interpolation:
    \begin{equation}
        \text{SS} = 1 - \frac{\text{CRPS}_{\text{model}}}{\text{CRPS}_{\text{ref}}}
        \label{eq:skill_score}
    \end{equation}
    where the reference CRPS equals the MAE of the bilinear baseline (since the CRPS of a deterministic forecast reduces to its mean absolute error). A skill score of~0 indicates parity with the baseline; positive values indicate improvement.

    \item[Correlations.] Pearson correlations between predicted uncertainty~($\sigma$) and absolute error, between altitude and error, between mTPI and error, and between day-of-year and error.

    \item[Calibration.] Probability Integral Transform (PIT) mean and standard deviation, observed coverage at the 50\%, 90\%, and 95\% nominal levels, the Kolmogorov--Smirnov $p$-value, and the Mean Absolute Calibration Error (MACE), to track the reliability of the uncertainty estimates.
\end{description}

\subsection{Predictions on sparse gridded input}

For gridded input, a sparse grid of ERA5-Land was offered the model as input to predict the full target grid. Different sparsity levels were captured.

\subsection{Predictions on sparse out-of-grid input}

For out-of-grid input (sampled from the MeteoSwiss target dataset), the points are regridded onto the training grid (ERA5-Land) and then fed to the model for full target grid prediction. Given the rugged Swiss topography, this approach is expected to cause a massive reduction in model performance.

Please see section \ref{sec:out-of-grid} in the Appendix for considerations on how to better handle out-of-grid inputs.

\subsection{Technology Stack}

The project is implemented in Python on the following stack:

\begin{itemize}
    \item \textbf{PyTorch} for model definition, training, and GPU inference.
    \item \textbf{xarray} and \textbf{Dask} for lazy-loading and processing of NetCDF/Zarr datasets.
    \item \textbf{pyproj} for WGS84\,/\,LV95 coordinate transformations.
    \item \textbf{scipy} and \textbf{scikit-learn} for evaluation metrics and preprocessing.
    \item \textbf{Renku} \citep{renku} (SDSC) for reproducible environments, GPU cloud sessions (NVIDIA MIG partitions), and version control.
\end{itemize}

Reproducibility is enforced by seeding all random number generators (Python, NumPy, PyTorch, CUDA) with a fixed seed and setting \texttt{cudnn.deterministic=True}.
\chapter{Results and Evaluation}

\section{Experimental Results}
\label{sec:experimental_results}

This section demonstrates that the ConvCNP training procedure converges reliably and that the resulting model produces physically plausible downscaled predictions. Detailed quantitative evaluation follows in subsequent sections.

\subsection{Training Convergence}

\begin{figure}[H]
    \centering
    \includegraphics[width=\textwidth]{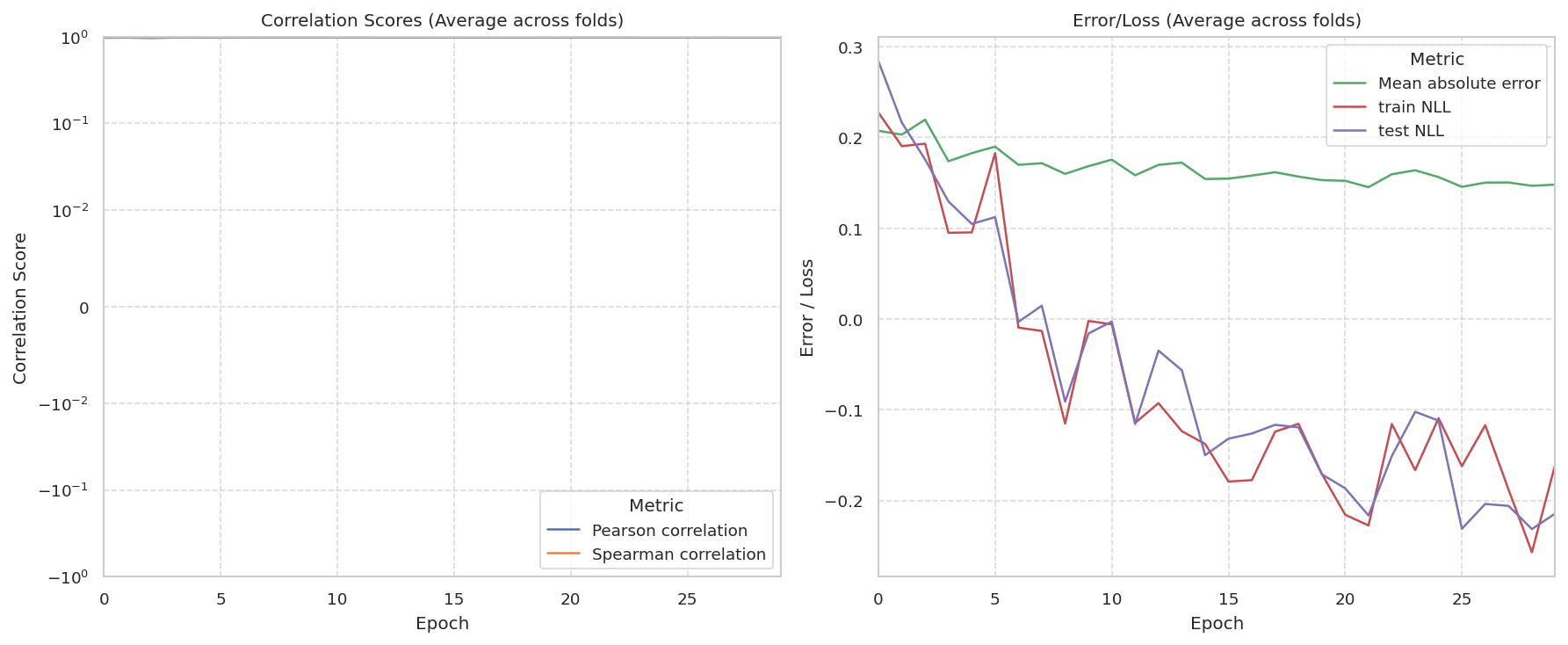}
    \caption{Training curves for \texttt{final-2014-30e-5f}, averaged across five folds. Left: Pearson and Spearman correlations (log residual scale). Right: validation MAE (normalized units), training NLL, and validation NLL.}
    \label{fig:training_curves}
\end{figure}

Figure~\ref{fig:training_curves} shows the training curves for the primary ten-year model \newline (\texttt{final-2014-30e-5f}), averaged across all five cross-validation folds. The left panel tracks Pearson and Spearman correlation on a logarithmic residual scale ($1 - r$); the right panel shows the validation MAE (in normalized units) alongside the training and validation negative log-likelihood (NLL).

Three convergence trends are evident:

\begin{enumerate}
    \item \textbf{Error reduction.} The validation MAE drops steeply during the first five epochs---from ${\sim}0.21$ to ${\sim}0.16$ in normalized units---and continues to decrease more gradually thereafter, reaching ${\sim}0.14$ by epoch~30. All five folds exhibit the same pattern with tight variance: the best per-fold MAE values cluster around $0.141 \pm 0.001$, indicating consistent convergence.

    \item \textbf{Correlation improvement.} Both Pearson and Spearman correlations start high---above 0.967 already at epoch~0, reflecting the strong spatial structure in the temperature field---and improve to ${\sim}0.98$ by the end of training.

    \item \textbf{Loss decrease.} The training and validation NLL both decrease from positive values (${\sim}0.2$) to negative values (${\sim}{-}0.25$), indicating that the model learns to assign increasingly high likelihood to the observed targets. The two curves track each other closely throughout, with no divergence between them---a sign that overfitting is not occurring. No fold triggered early stopping; all five ran for the full 30~epochs, and the best checkpoints were selected between epochs~18 and~29, suggesting that marginal gains were still accumulating toward the end of training. This is further confirmed by the experiment running for up to 100 epochs (see below).
\end{enumerate}

\subsection{Sample Prediction}

\begin{figure}[H]
    \centering
    \includegraphics[width=\textwidth]{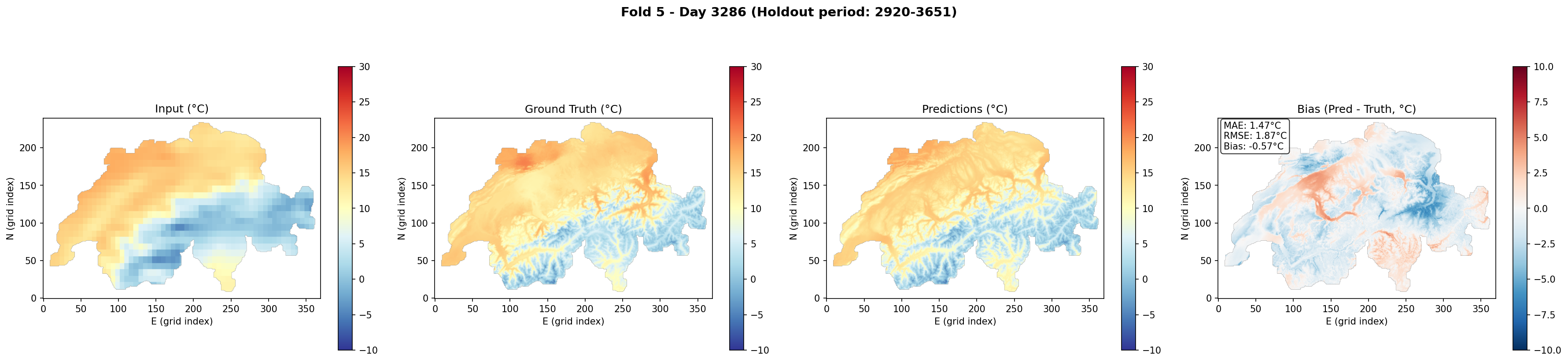}
    \caption{Single-day prediction for fold~5 (day~3\,286) of the ten-year model. From left to right: ERA5-Land input (${\sim}$11\,km), MeteoSwiss ground truth (${\sim}$1\,km), model prediction (${\sim}$1\,km), and bias (prediction~$-$~truth). For this day and fold, MAE~=~1.47\,\textdegree C, RMSE~=~1.87\,\textdegree C, bias~=~$-$0.57\,\textdegree C.}
    \label{fig:sample_prediction}
\end{figure}

Figure~\ref{fig:sample_prediction} illustrates a single-day prediction from the fifth cross-validation fold.

The input panel shows the smooth, coarse-resolution temperature field from ERA5-Land, in which topographic detail is largely absent. The ground-truth panel reveals the fine-grained spatial structure that the model must recover: sharp temperature gradients along the Alpine ridge, cold mountain peaks, and warmer valleys---features well below the resolution of the input grid.

The prediction panel demonstrates that the model successfully reconstructs much of this detail. The overall spatial pattern closely follows the ground truth, with the major valleys, plateaus, and high-altitude regions clearly resolved. The bias panel confirms that residual errors are small across most of the domain (near-white shading), with the largest deviations concentrated in high-mountain areas where the elevation gradient is steepest.

\section{Performance}
\subsection{Global metrics}

Here are visualizations of the errors, error bias, CRPS and skill for the baseline model (10 years of data, 30 epochs, 5 folds, no ablation, aggregate results over the folds).

\begin{figure}[H]
    \centering
    \includegraphics[width=\textwidth, trim={0 0 0 20px}, clip]{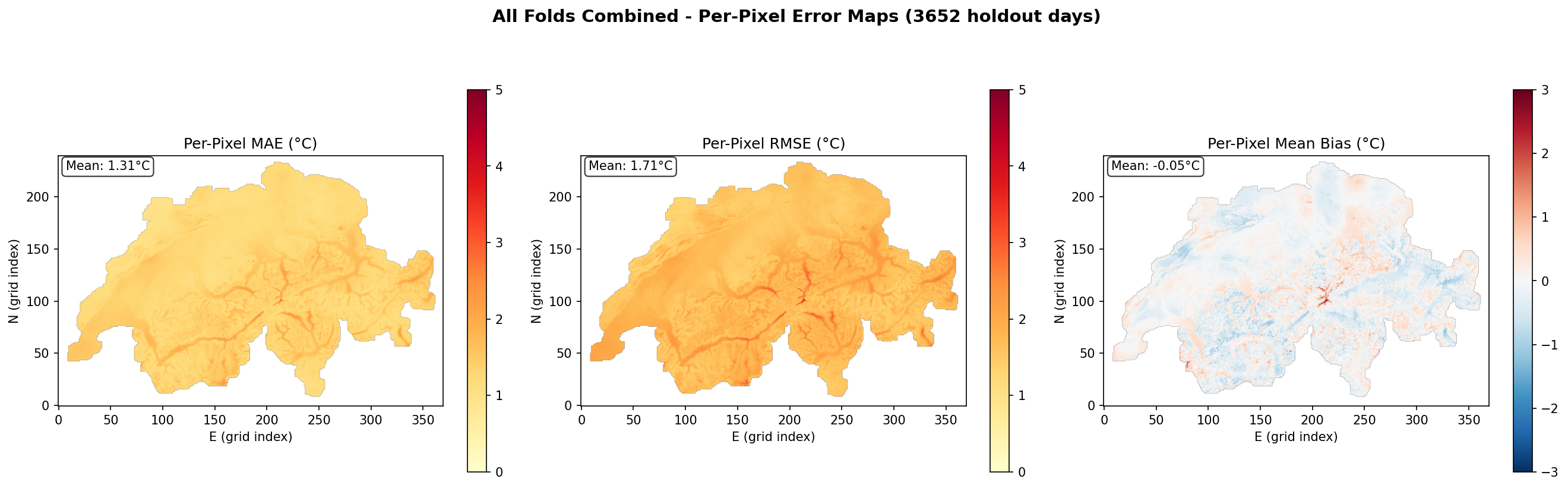}
    \caption{Per-pixel MAE, RMSE and error bias on maps}
    \label{fig:joint_error_maps}
\end{figure}

\begin{figure}[H]
    \centering
    \begin{tabular}{@{}cc@{}}
        \includegraphics[width=0.48\textwidth, trim={0 0 0 25px}, clip]{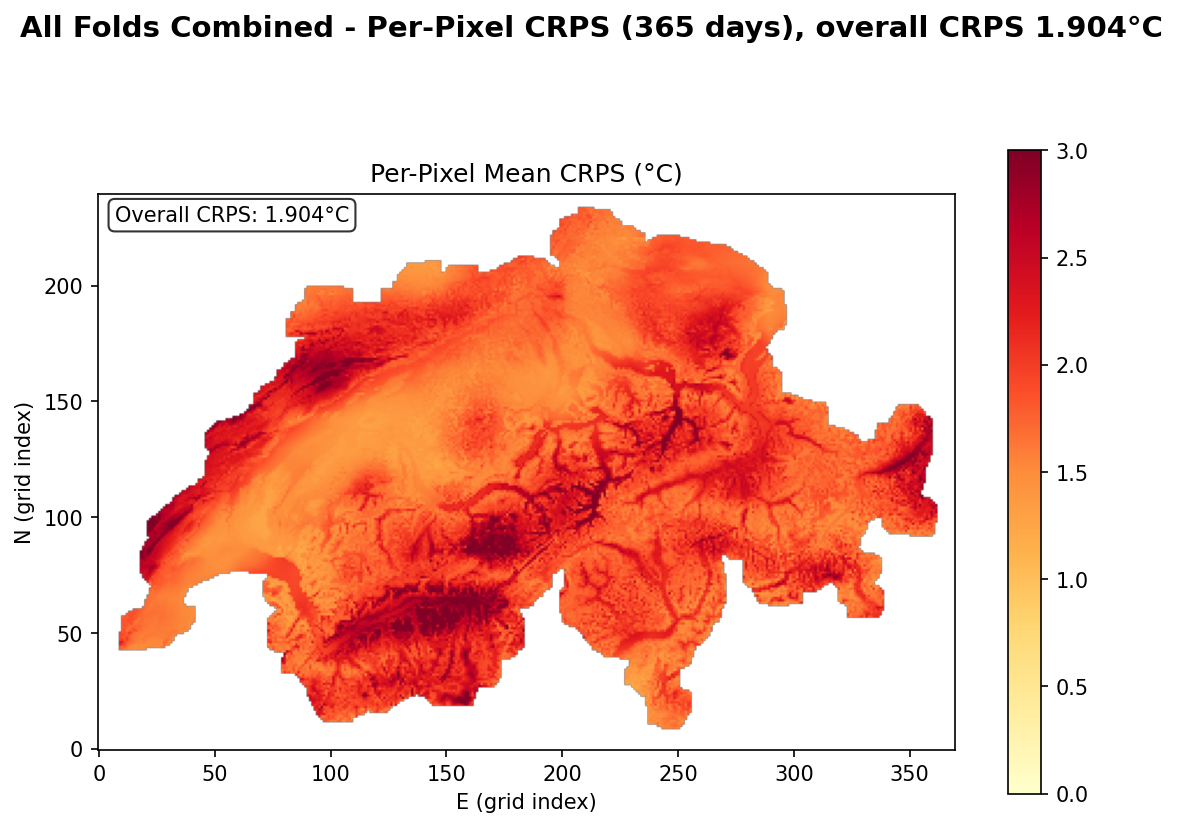} &
        \includegraphics[width=0.48\textwidth, trim={0 0 0 25px}, clip]{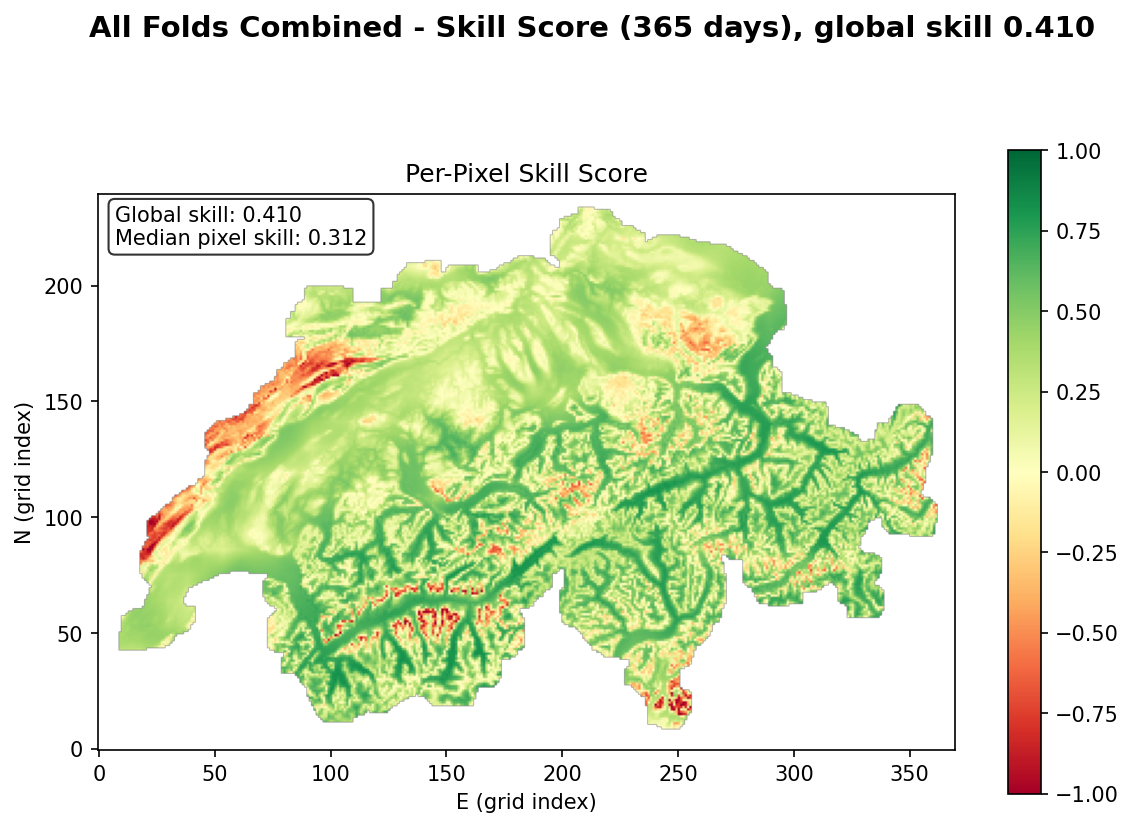} \\
    \end{tabular}
    \caption{Per-pixel CRPS and Skill maps}
    \label{fig:crps_skill_maps}
\end{figure}

\begin{table}[H]
\centering
\caption{Full-grid evaluation overview. All models evaluated with 100\% of the ERA5-Land grid (1\,769~points) as context input.}
\label{tab:full_grid_overview}
\begin{tabular}{@{}lrrrrr@{}}
\toprule
\textbf{Model} & \textbf{MAE} & \textbf{RMSE} & \textbf{Bias} & \textbf{CRPS} & \textbf{Skill} \\
                & (\textdegree C) & (\textdegree C) & (\textdegree C) & (\textdegree C) & \\
\midrule
\multicolumn{6}{@{}l}{\textit{Primary models}} \\
\quad \texttt{2023-10e-5f}   & 2.09 & 2.67 & $+$0.66 & 1.90 & 0.245 \\
\quad \texttt{2023-30e-5f}   & 1.64 & 2.08 & $+$0.24 & 1.49 & 0.410 \\
\quad \texttt{2023-100e-5f}  & 1.62 & 2.05 & $+$0.25 & 1.48 & 0.415 \\
\quad \texttt{2014-30e-5f}  (base model)   & 1.31 & 1.71 & $-$0.05 & 1.21 & 0.524 \\
\midrule
\bottomrule
\end{tabular}
\end{table}

\newpage
Table~\ref{tab:full_grid_overview} summarises the headline accuracy and skill metrics for all evaluated models on the full ERA5-Land grid (1\,769 context points). Each row reports the mean absolute error (MAE), root-mean-square error (RMSE), and mean bias of the point prediction in degrees Celsius, together with the Continuous Ranked Probability Score (CRPS) and a skill score defined as $\text{Skill} = 1 - \text{CRPS}_{\text{model}} / \text{CRPS}_{\text{ref}}$, where the reference is the deterministic bilinear interpolation of ERA5-Land to the target grid (for a deterministic forecast, $\text{CRPS} = \text{MAE}$). A skill of~0 indicates no improvement over the interpolation baseline; a skill of~1 would indicate a perfect forecast.

\paragraph{Effect of training duration.}
The three single-year models (trained on 2023 data, evaluated on 365 holdout days) isolate the effect of longer optimisation. Increasing from 10 to 30~epochs cuts the MAE from 2.09 to 1.64\,\textdegree C (a 22\% reduction) and nearly doubles the skill score from 0.245 to 0.410. Extending training further to 100~epochs yields only a marginal additional gain (MAE 1.62\,\textdegree C, Skill 0.415), suggesting that most of the representational capacity available to the single-year model is utilised within the first 30~epochs. Even though the slowdown was found, up to 100 epochs convergence continues without signs of overfitting.

\paragraph{Effect of training data volume.}
Moving from one year to ten years of training data (both at 30~epochs) produces the largest single improvement: the MAE drops from 1.64 to 1.31\,\textdegree C, the RMSE from 2.08 to 1.71\,\textdegree C, and the skill score rises from 0.410 to 0.524. The mean bias also shrinks to near zero ($-$0.05\,\textdegree C). These gains indicate that additional climatological diversity---exposing the model to a wider range of synoptic situations and seasonal extremes---is more beneficial than additional gradient steps on a limited sample.

\paragraph{Skill relative to ERA5-Land interpolation.}
All primary models improve upon the bilinear-interpolation baseline. The ten-year model achieves a CRPS-based skill of 0.524, meaning its probabilistic predictions reduce the expected error by more than half compared to simply interpolating the coarse reanalysis field. Even the weakest model (10~epochs, single year) attains a skill of 0.245, confirming that the ConvCNP learns meaningful fine-scale structure beyond what spatial interpolation alone can provide.

\paragraph{Uncertainty calibration.}
\nopagebreak
\begin{figure}[H]
    \centering
    \begin{tabular}{@{}cc@{}}
        \includegraphics[width=0.48\textwidth]{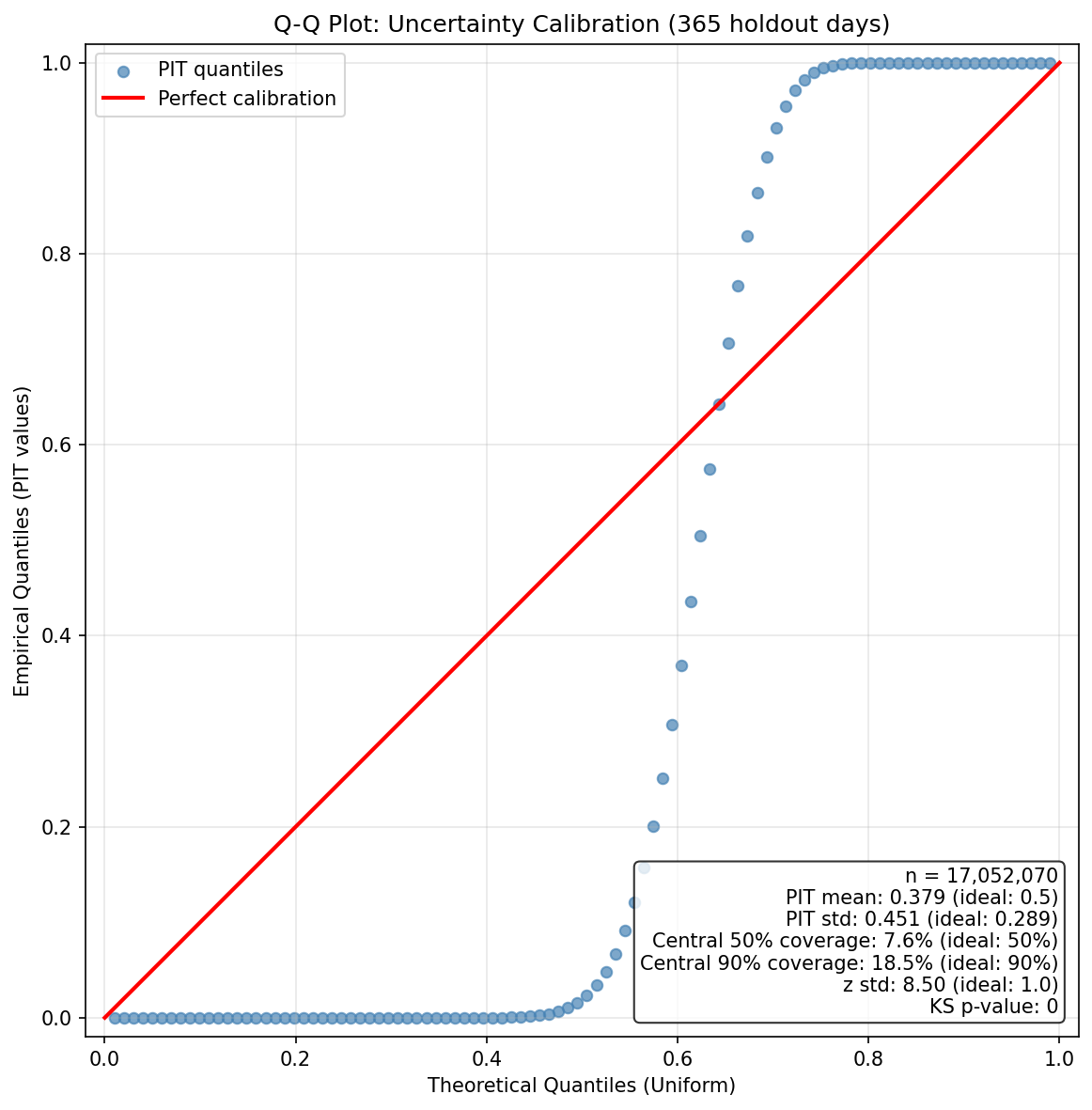} &
        \includegraphics[width=0.48\textwidth]{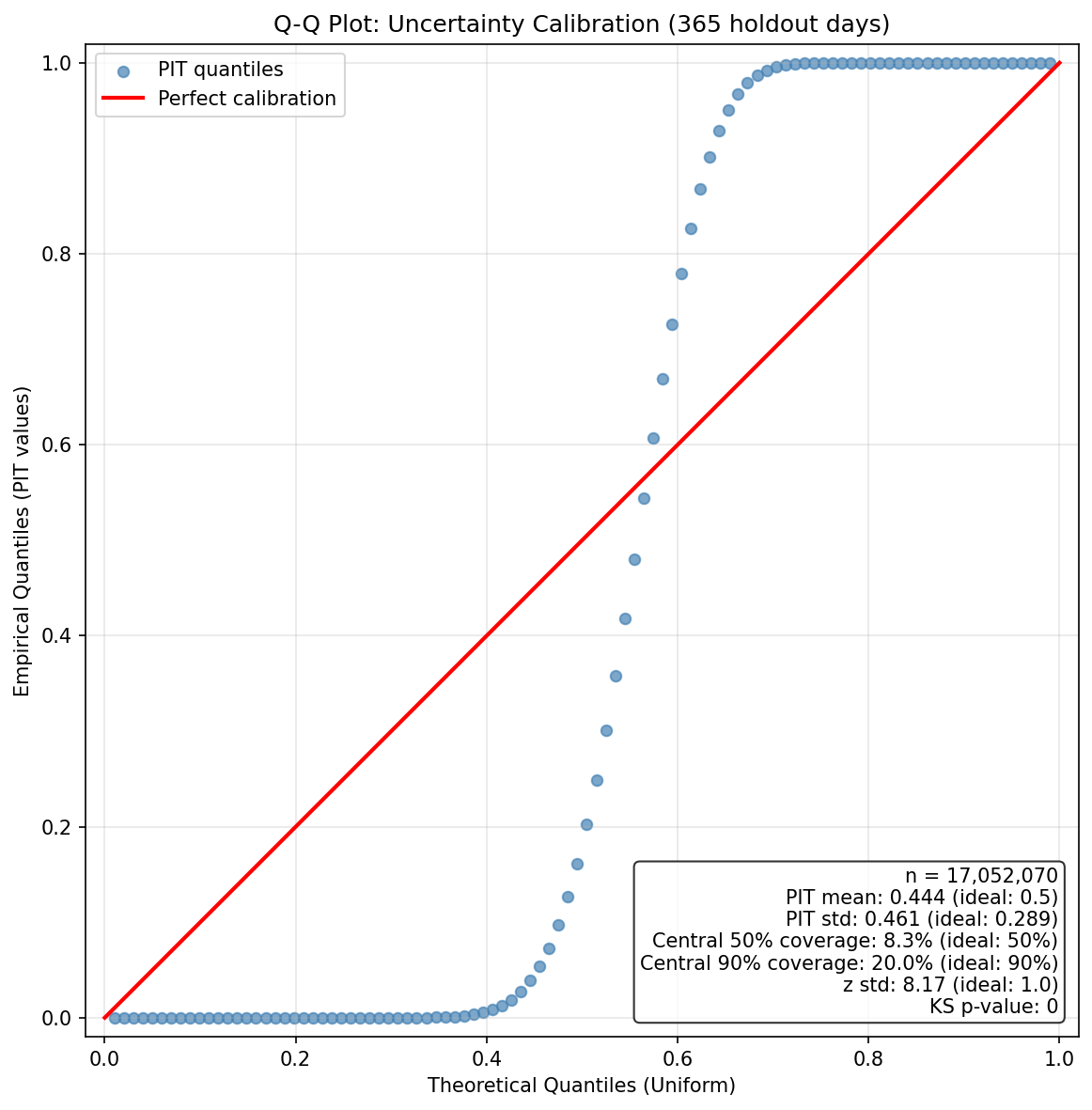} \\
        \includegraphics[width=0.48\textwidth]{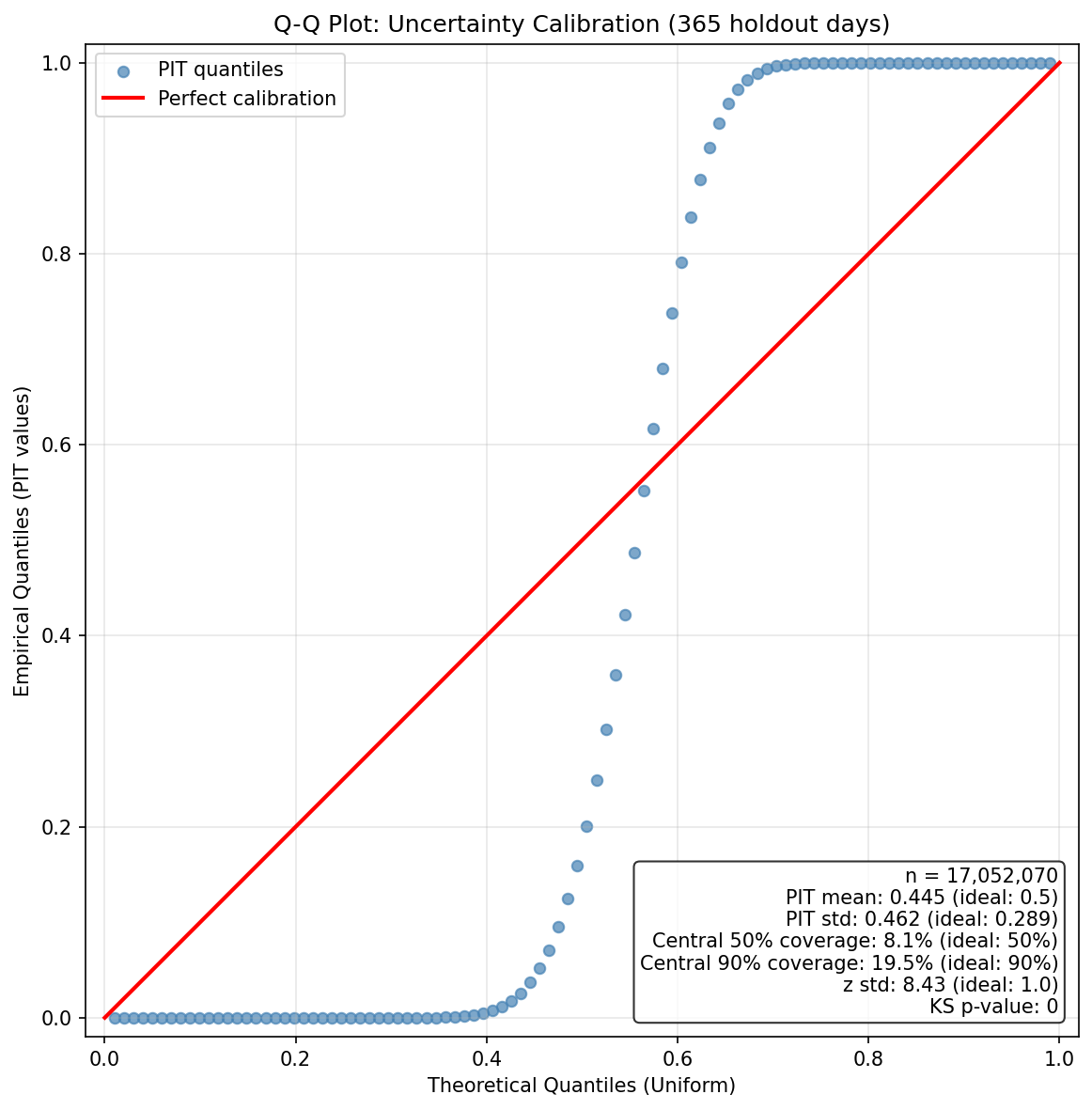} &
        \includegraphics[width=0.48\textwidth]{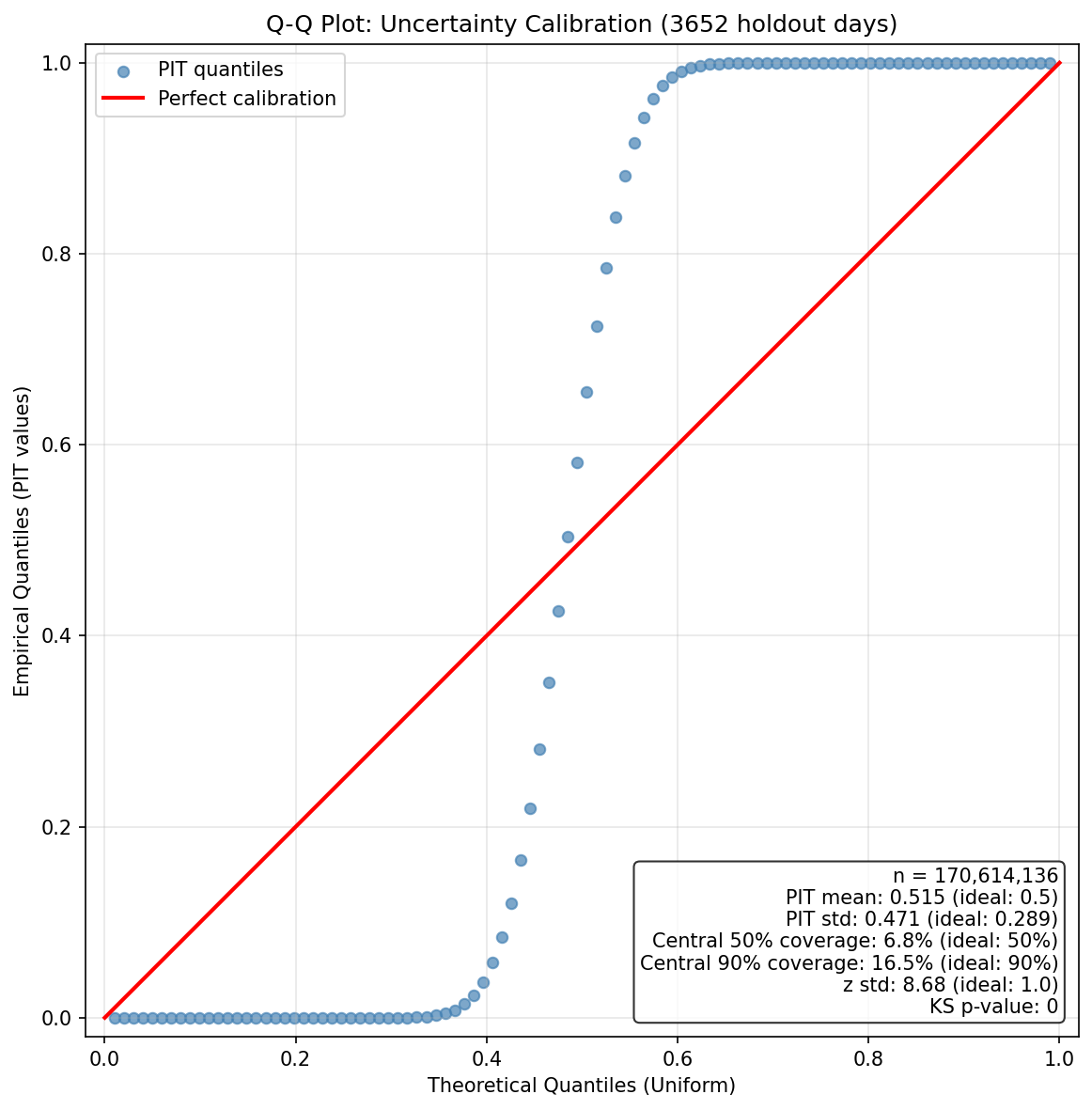} \\
    \end{tabular}
    \caption{Q--Q calibration plots (PIT quantiles vs.\ uniform theoretical quantiles) for the four primary models. Top row: single-year models at 10 and 30~epochs. Bottom row: single-year model at 100~epochs (left) and ten-year model at 30~epochs (right). All models are substantially overconfident, with central 90\% coverage ranging from 16.5\% to 20\%.}
    \label{fig:qq_calibration}
\end{figure}

Figure~\ref{fig:qq_calibration} shows Q--Q plots of the Probability Integral Transform (PIT) for each primary model. A perfectly calibrated model would produce PIT values uniformly distributed on $[0,1]$, yielding points along the diagonal. All four models exhibit a pronounced S-shaped departure: the empirical quantiles rise steeply, the standardised residual standard deviation ($z$-std) is approximately 8.7 (ideal: 1.0), confirming that the model is overconfident by roughly an order of magnitude.

This miscalibration is consistent across all configurations and may arise from the Gaussian likelihood training objective, which optimises overall fit rather than interval coverage. 

\subsection{Correlations}

\begin{figure}[H]
    \centering
    \includegraphics[width=\textwidth, trim={0 0 0 25px}, clip]{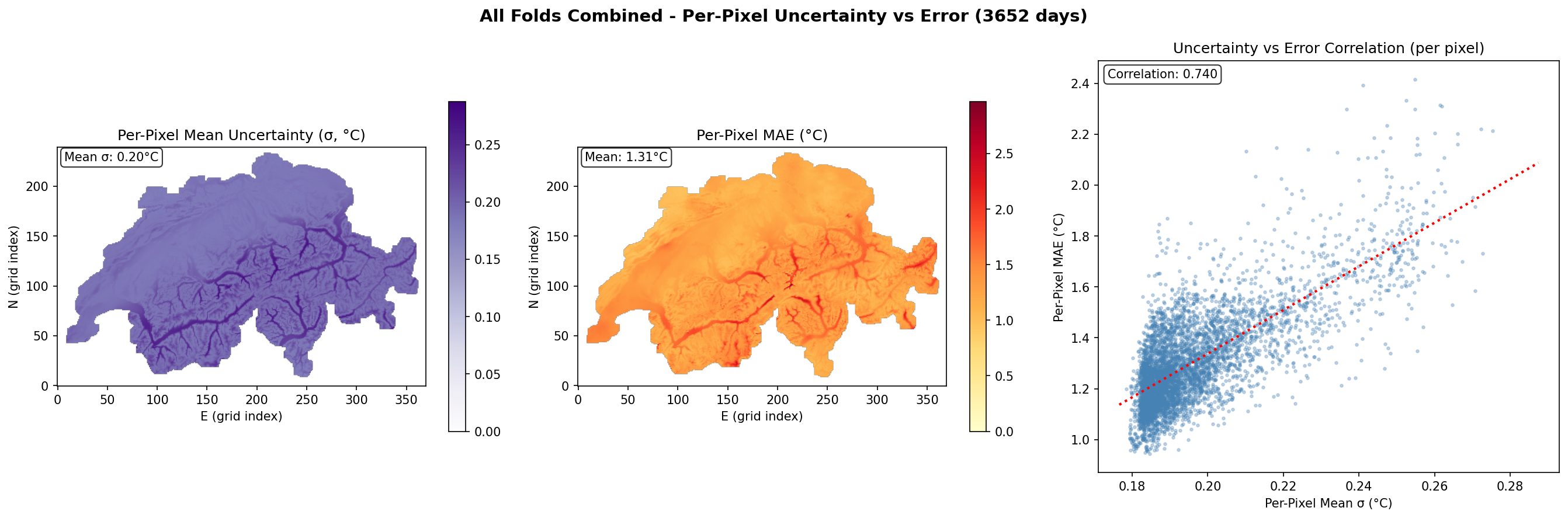}
    \caption{Uncertainty map, error map for comparison, and cross-correlation on the baseline model, covering 10 years of training data, 30 epochs, 5 folds, no ablation, conjoined results from all folds.}
    \label{fig:uncertainty_vs_error}
\end{figure}

Despite the overconfidence mentioned above, the model's predicted uncertainties remain informative: the correlation between predicted $\sigma$ and absolute error is 0.74 for the ten-year model, meaning that the uncertainty estimates reliably rank predictions by quality even though their absolute magnitudes are too small.

\begin{figure}[H]
    \centering
    \includegraphics[width=\textwidth, trim={0 0 0 25px}, clip]{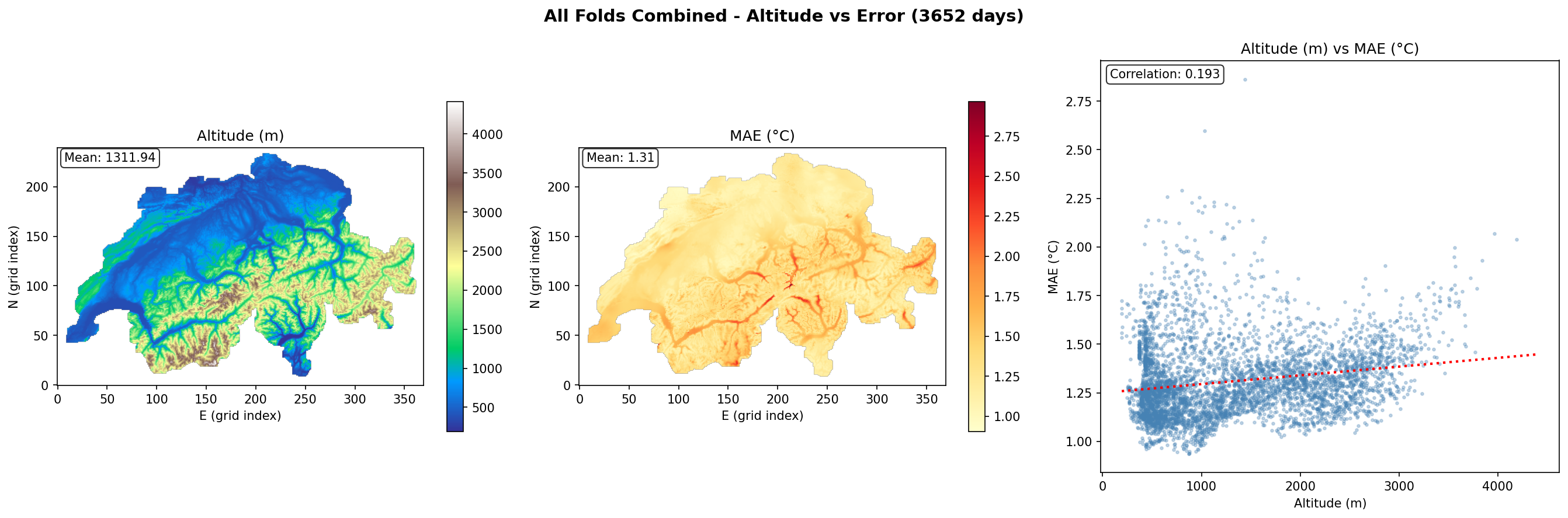}
    \caption{Altitude map, error map for comparison, and cross-correlation on the baseline model, covering 10 years of training data, 30 epochs, 5 folds, no ablation, conjoined results from all folds.}
    \label{fig:altitude_vs_error}
\end{figure}

It is known that weather prediction at altitude is a tough problem. The correlation between altitude and error exists, but is significantly lower than the one observed between predicted uncertainty and error.

\begin{figure}[H]
    \centering
    \includegraphics[width=\textwidth, trim={0 0 0 25px}, clip]{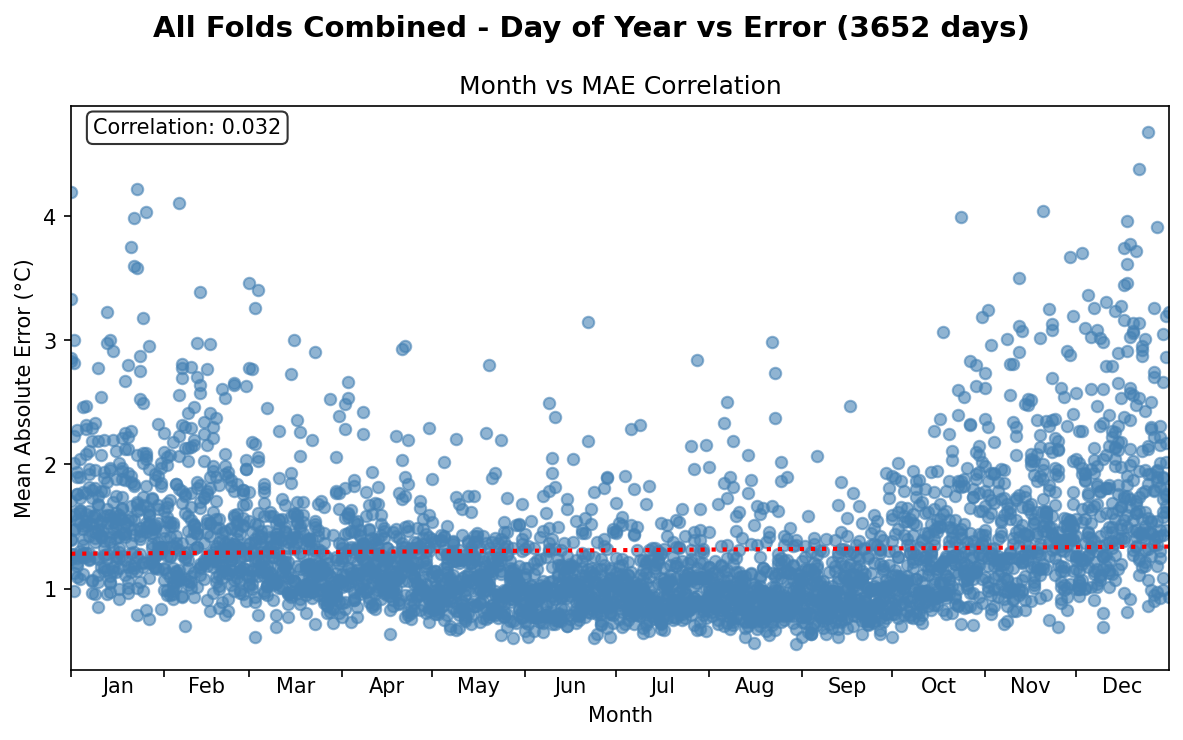}
    \caption{Mean error per day-of-year, covering 10 years of training data, 30 epochs, 5 folds, no ablation, conjoined results from all folds.}
    \label{fig:doy_vs_error}
\end{figure}

We can observe the colder months (November--March) observe a higher incidence of outliers in mean absolute error.

\section{Ablation Analysis}
\label{sec:ablation}

\begin{table}[H]
\centering
\caption{Ablation evaluation overview. All models evaluated with 100\% of the ERA5-Land grid (1\,769~points) as context input.}
\label{tab:ablation_overview}
\begin{tabular}{@{}lrrrrr@{}}
\toprule
\textbf{Model} & \textbf{MAE} & \textbf{RMSE} & \textbf{Bias} & \textbf{CRPS} & \textbf{Skill} \\
                & (\textdegree C) & (\textdegree C) & (\textdegree C) & (\textdegree C) & \\
\midrule
\multicolumn{6}{@{}l}{\textit{Primary model (2014--2023)}} \\
\quad \texttt{2014-30e-5f}  (base model)   & 1.31 & 1.71 & $-$0.05 & 1.21 & 0.524 \\
\midrule
\multicolumn{6}{@{}l}{\textit{Ablation models (2014--2023)}} \\
\quad \texttt{-ablation-elevation}\textsuperscript{$\dagger$}
                              & \multicolumn{5}{c}{--- model diverged; see text ---} \\
\quad \texttt{-ablation-mtpi}            & 1.33 & 1.74 & $-$0.13 & 1.23 & 0.517 \\
\quad \texttt{-ablation-seasonal-all}    & 1.38 & 1.80 & $-$0.12 & 1.27 & 0.50 \\
\quad \texttt{-ablation-seasonal-in-mlp} & 1.29 & 1.69 & $+$0.08 & 1.19 & 0.532 \\
\bottomrule
\end{tabular}
\begin{flushleft}
\textsuperscript{$\dagger$}\small Without the elevation MLP, the model produced errors orders of magnitude above physical plausibility (MAE~$\approx$~802\,\textdegree C), indicating that the topographic correction stage is essential for meaningful predictions. (This is unexpected, we would expect some degree of interpolation to still take place.) \\[2pt]
\end{flushleft}
\end{table}

All ablation models use the same ten-year dataset (2014--2023), 30 training epochs, five-fold cross-validation, and identical splits as the base model. Each experiment disables a single feature group while keeping everything else unchanged, so that differences in the metrics can be attributed to that component alone.

\paragraph{Elevation features.}
Removing the three topographic inputs---true elevation, elevation difference, and TPI---from the elevation MLP (\texttt{USE\_ELEVATION=False}) caused the model to diverge catastrophically. The resulting MAE of ${\sim}802$\,\textdegree C and a negative skill score of $-314$ indicate that the predictions are off-scale by orders of magnitude. Notably, the altitude--error correlation jumps from 0.19 in the base model to 0.78 in this ablation, confirming that the errors are overwhelmingly concentrated at locations where the true elevation differs most from the coarse ERA5-Land grid-cell elevation. Without topographic correction, the elevation MLP receives only the CNN's interpolated $(\mu, \sigma)$ and zero-valued placeholders; it has no mechanism to compensate for the fact that a 3\,000\,m peak and a 400\,m valley within the same ${\sim}$11\,km grid cell receive nearly identical CNN predictions. The MLP, trained on these uninformative inputs, learns weight configurations that produce numerically unstable outputs at evaluation time---when all 46\,000+ target points are predicted simultaneously---resulting in the observed divergence. This outcome confirms that the elevation MLP is not merely a refinement stage but the \emph{essential} component that bridges the resolution gap between the coarse reanalysis grid and the fine-scale target field. The author must also consider that there may have been some misconfiguration or bug which could cause this, but one was not identified.

\paragraph{Topographic Position Index.}
Removing only the TPI channel \newline (\texttt{USE\_MTPI=False}) while retaining true elevation and elevation difference has a marginal effect: the MAE increases from 1.31 to 1.33\,\textdegree C and the skill score decreases from 0.524 to 0.517. The bias shifts from $-0.05$ to $-0.13$\,\textdegree C, a small systematic cold bias. This suggests that TPI---which encodes whether a target lies on a ridge, in a valley, or on an open slope---provides a secondary, but not essential, correction on top of the elevation difference. The elevation difference alone captures most of the sub-grid topographic variance.

\paragraph{Seasonal features.}
Two ablations target the cyclical day-of-year encoding ($\cos$/$\sin$ channels). The first (\texttt{-ablation-seasonal-all}) removes them everywhere: both from the CNN input channels and from the elevation MLP. We can see fully ablating time-of-year encoding (both in the spatially-aware CNN and the MLP) has a significant impact on model performance.

\paragraph{Seasonal feature ablation from the MLP outperforms the base model.}
Counter-intuitively, this ablation achieve slightly \emph{better} metrics than the base model (MAE 1.29 vs.\ 1.31\,\textdegree C; Skill 0.532 vs.\ 0.524). The improvement is small but consistent across all metrics. This likely reflects a mild regularisation effect: the base model's elevation MLP has 7~inputs (including $\cos$/$\sin$), while the seasonal ablations have only 5, reducing the MLP's capacity to overfit to spurious correlations between the seasonal encoding and the residual error. Alternatively, feeding redundant features to the MLP may slightly impede optimisation by diluting the gradient signal from the more informative elevation inputs. In either case, the practical difference is modest, and the result suggests that removing the explicit seasonal features is, if anything, slightly beneficial. This also suggests the MLP might be constrained and might capture more differentiable signal from the training data if we increase its size.

\section{Performance on sparse input, on- and off-grid}
\label{sec:sparse}

A key question for real-world deployment is how the model behaves when the full ERA5-Land grid is unavailable and only a sparse set of observations is provided as context. We evaluate two scenarios: (i)~\emph{on-grid} sparsity, where a random subset of ERA5-Land grid cells is retained, and (ii)~\emph{off-grid} sparsity, where the context consists of MeteoSwiss ground-station observations placed onto the nearest ERA5 grid cell. In both cases the model weights are frozen---no retraining is performed---so the results measure zero-shot robustness to reduced input.

\subsection{On-grid sparsity (ERA5-Land)}

\begin{table}[H]
\centering
\caption{Performance with sparse ERA5-Land grid input. The full grid has 1\,769~points.}
\label{tab:sparse_era5}
\begin{tabular}{@{}lrrrrrrr@{}}
\toprule
\textbf{Input} & \textbf{Points} & \textbf{MAE} & \textbf{RMSE} & \textbf{Bias} & \textbf{CRPS} & \textbf{Skill} & \textbf{$\sigma$--err} \\
                & & (\textdegree C) & (\textdegree C) & (\textdegree C) & (\textdegree C) & & corr \\
\midrule
\quad 20 pts      & 20   & 4.93 & 6.09 & $+$1.81 & 4.83 & $-$0.903 & 0.111 \\
\quad 50 pts      & 50   & 3.73 & 4.66 & $+$1.73 & 3.63 & $-$0.430 & $-$0.012 \\
\quad 80 pts      & 80   & 3.20 & 4.02 & $+$1.69 & 3.10 & $-$0.223 & $-$0.023 \\
\quad 100 pts     & 100  & 2.99 & 3.75 & $+$1.59 & 2.89 & $-$0.140 & $-$0.026 \\
\quad 120 pts     & 120  & 2.77 & 3.48 & $+$1.43 & 2.67 & $-$0.052 & $-$0.010 \\
\quad 10\% (177)  & 177  & 2.27 & 2.86 & $+$1.01 & 2.17 & 0.144  & $-$0.002 \\
\quad 25\% (442)  & 442  & 1.51 & 1.93 & $+$0.20 & 1.41 & 0.446  & 0.272 \\
\quad 50\% (885)  & 885  & 1.36 & 1.77 & $+$0.01 & 1.26 & 0.505  & 0.668 \\
\quad 90\% (1\,592) & 1\,592 & 1.31 & 1.72 & $-$0.04 & 1.21 & 0.522 & 0.736 \\
\quad 100\% (1\,769) & 1\,769 & 1.31 & 1.71 & $-$0.05 & 1.21 & 0.524 & 0.740 \\
\bottomrule
\end{tabular}
\end{table}

Table~\ref{tab:sparse_era5} summarises the model's performance as the fraction of retained ERA5-Land grid cells decreases from 100\% to as few as 20~points.

\paragraph{Graceful degradation at moderate sparsity.}
Halving the input grid to 50\% (885~points) increases the MAE by only 3.8\%, from 1.31 to 1.36\,\textdegree C, and the skill score drops marginally from 0.524 to 0.505. At 25\% of the grid (442~points), the MAE rises to 1.51\,\textdegree C (+15\%) but the skill remains strongly positive at 0.446, meaning the model still substantially outperforms bilinear interpolation. These results demonstrate that the ConvCNP's convolutional encoder can extract useful spatial structure even from heavily subsampled gridded input.

\paragraph{Transition to negative skill.}
At 10\% of the grid (177~points), the MAE increases to 2.27\,\textdegree C (+74\%) and the skill drops to 0.144---still positive, but the model's advantage over the interpolation baseline is diminishing rapidly. Below 177~points, the skill turns negative: at 120~points it is $-$0.052 and at 20~points it reaches $-$0.903, indicating that the model performs worse than simple bilinear interpolation. The crossover occurs between approximately 120 and 177~ERA5 grid points.

\paragraph{Uncertainty estimates collapse.}
At the full grid, the correlation between predicted $\sigma$ and absolute error is 0.74, indicating highly informative uncertainty estimates. This correlation degrades smoothly---0.67 at 50\%, 0.27 at 25\%---and effectively vanishes below 10\% ($-$0.002 at 177~points). For the absolute-count configurations (20--120~points), the $\sigma$--error correlation hovers near zero, meaning the model's uncertainty estimates are no longer informative. This is consistent with the CNN receiving too little spatial context to distinguish uncertain from confident predictions.

\paragraph{Systematic warm bias.}
At full coverage the model exhibits near-zero bias ($-$0.05\,\textdegree C). As context is removed, a systematic warm bias emerges: $+$0.20\,\textdegree C at 25\%, $+$1.01\,\textdegree C at 10\%, and $+$1.81\,\textdegree C at 20~points. When the CNN receives too few observations to reconstruct the spatial temperature pattern, the elevation MLP's residual correction---trained to compensate for the difference between the coarse ERA5 grid-cell temperature and the fine-scale target---reverts toward a warm-biased default.

\begin{figure}[H]
    \centering
    \includegraphics[width=\textwidth]{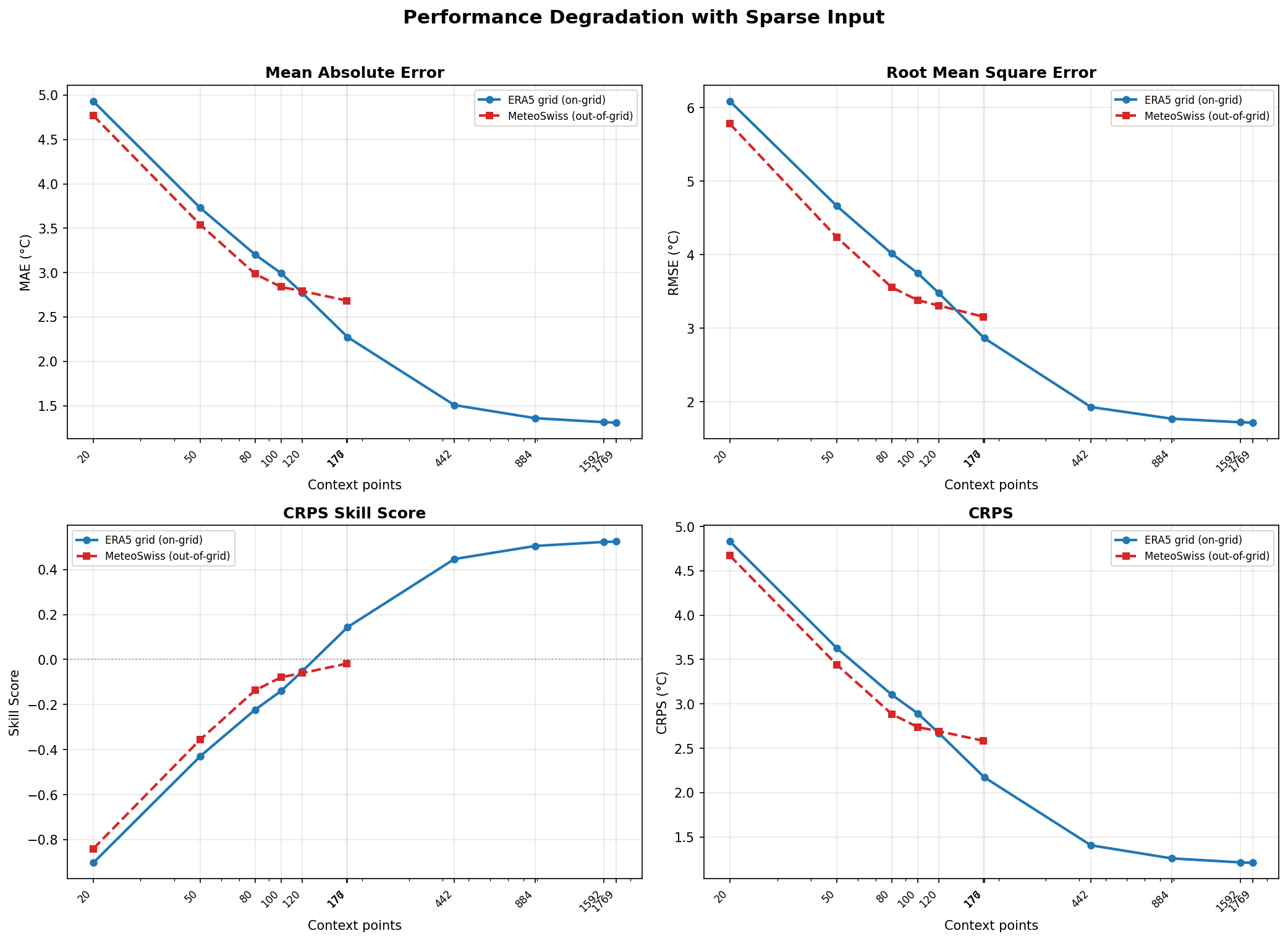}
    \caption{MAE, RMSE, CRPS, and Skill vs.\ number of context points for ERA5 on-grid (blue circles) and MeteoSwiss off-grid (red squares) input.}
    \label{fig:metrics_vs_density}
\end{figure}

Figure~\ref{fig:metrics_vs_density} shows the degradation curves for both input types. The ERA5 on-grid curve exhibits a clear elbow around 25--50\% of the grid: above this range the model is near-optimal; below it, performance degrades rapidly.

\subsection{Off-grid sparsity (MeteoSwiss target set)}

\begin{table}[H]
\centering
\caption{Performance with sparse MeteoSwiss (off-grid) input. Maximum used: 176~points.}
\label{tab:sparse_meteoswiss}
\begin{tabular}{@{}lrrrrrrr@{}}
\toprule
\textbf{Input} & \textbf{Points} & \textbf{MAE} & \textbf{RMSE} & \textbf{Bias} & \textbf{CRPS} & \textbf{Skill} & \textbf{$\sigma$--err} \\
                & & (\textdegree C) & (\textdegree C) & (\textdegree C) & (\textdegree C) & & corr \\
\midrule
\quad 20 pts  & 20  & 4.77 & 5.78 & $+$2.05 & 4.67 & $-$0.840 & $-$0.047 \\
\quad 50 pts  & 50  & 3.54 & 4.24 & $+$2.26 & 3.44 & $-$0.356 & $-$0.178 \\
\quad 80 pts  & 80  & 2.99 & 3.56 & $+$1.98 & 2.89 & $-$0.137 & $-$0.135 \\
\quad 100 pts & 100 & 2.84 & 3.38 & $+$1.86 & 2.74 & $-$0.079 & $-$0.125 \\
\quad 120 pts & 120 & 2.79 & 3.31 & $+$1.92 & 2.69 & $-$0.060 & $-$0.128 \\
\quad 176 pts & 176 & 2.68 & 3.16 & $+$2.07 & 2.58 & $-$0.018 & $-$0.124 \\
\bottomrule
\end{tabular}
\end{table}

Table~\ref{tab:sparse_meteoswiss} presents results when the ERA5-Land grid is replaced entirely by MeteoSwiss metrics mapped onto the nearest ERA5 grid cell. This tests whether the model can operate from sparse, irregularly placed, off-grid measurements---a scenario closer to real-world station-based downscaling.

\paragraph{Negative skill at all densities.}
Even with all 176~available MeteoSwiss stations, the model achieves a skill of only $-$0.018---marginally worse than bilinear interpolation of the ERA5 field. At 20~stations the skill drops to $-$0.840. The model never outperforms the interpolation baseline in this configuration. Though this comparison is somewhat unfair: there is no real-time feed of a complete ERA5-Land grid; it only becomes available later after systemic and painstaking reanalysis.

\paragraph{Persistent warm bias.}
All MeteoSwiss configurations exhibit a large, stable warm bias between $+$1.86 and $+$2.26\,\textdegree C. Unlike the ERA5 on-grid case where the bias grows with sparsity, here the bias is approximately constant across all point counts. This suggests a systematic offset: MeteoSwiss stations, which are predominantly located in valleys and at lower elevations, provide a biased sample of the temperature field. The model, trained on full ERA5 grids, has no mechanism to correct for this sampling bias at inference time.

\paragraph{Uninformative uncertainty.}
The $\sigma$--error correlation is negative across all MeteoSwiss configurations ($-$0.05 to $-$0.18), meaning the model's predicted uncertainty is slightly \emph{anti}-correlated with the actual error. The model's uncertainty machinery, trained on dense gridded input, produces meaningless confidence estimates when the input distribution shifts to sparse off-grid points.

\subsection{On-grid vs.\ off-grid at comparable point counts}

\begin{figure}[H]
    \centering
    \includegraphics[width=\textwidth]{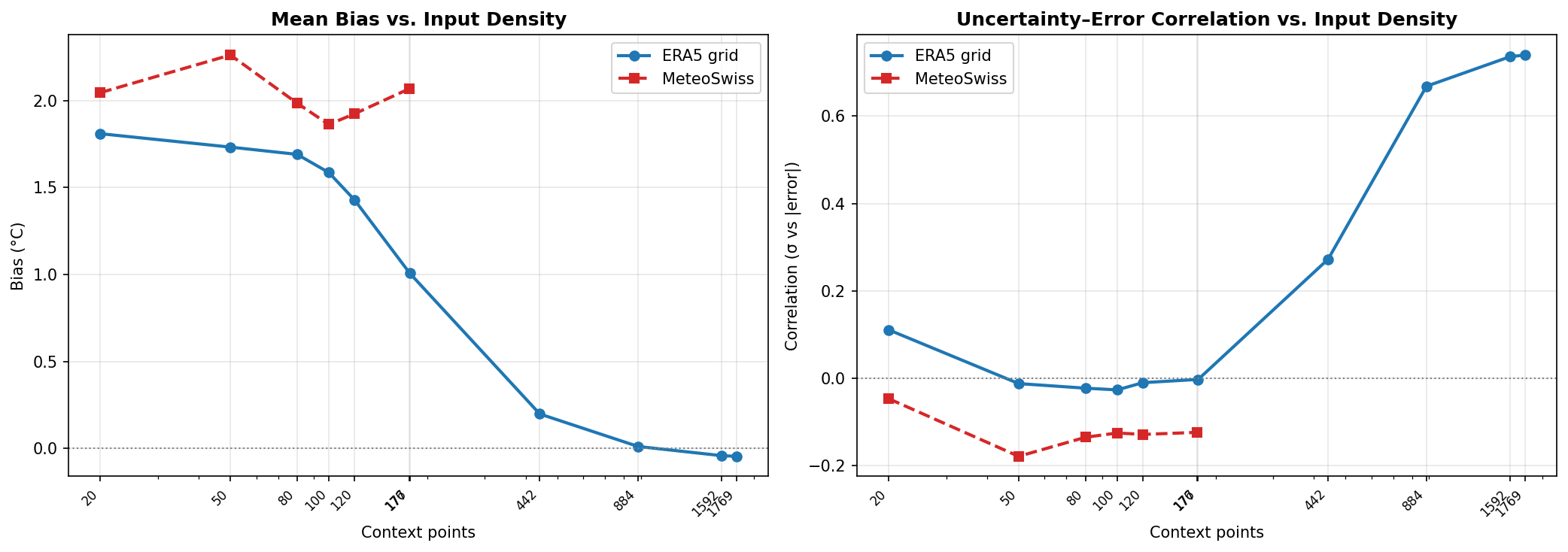}
    \caption{Left: mean bias vs.\ context density. The ERA5 bias grows from near-zero toward $+$1.8\,\textdegree C as points decrease; the MeteoSwiss bias remains around $+$2\,\textdegree C regardless of density. Right: $\sigma$--error correlation vs.\ density. On-grid prediction maintains informative uncertainty down to ${\sim}$25\% of the grid; off-grid uncertainty is never informative.}
    \label{fig:bias_uncertainty}
\end{figure}

At matched point counts (20, 50, 80, 100, 120), the ERA5 on-grid and MeteoSwiss off-grid configurations yield surprisingly similar MAE values---within 0.02--0.22\,\textdegree C of each other. At very low counts (20--80~points), MeteoSwiss actually achieves slightly lower MAE than ERA5 on-grid (e.g.\ 4.77 vs.\ 4.93\,\textdegree C at 20~points), likely a bias introduced because MeteoSwiss points are actually used as ground truth. However, this MAE advantage does not translate to positive skill, because the MeteoSwiss configurations suffer from a much larger systematic bias (${\sim}$2\,\textdegree C vs.\ ${\sim}$1.4--1.8\,\textdegree C for ERA5 at the same counts).

\begin{figure}[H]
    \centering
    \includegraphics[width=\textwidth]{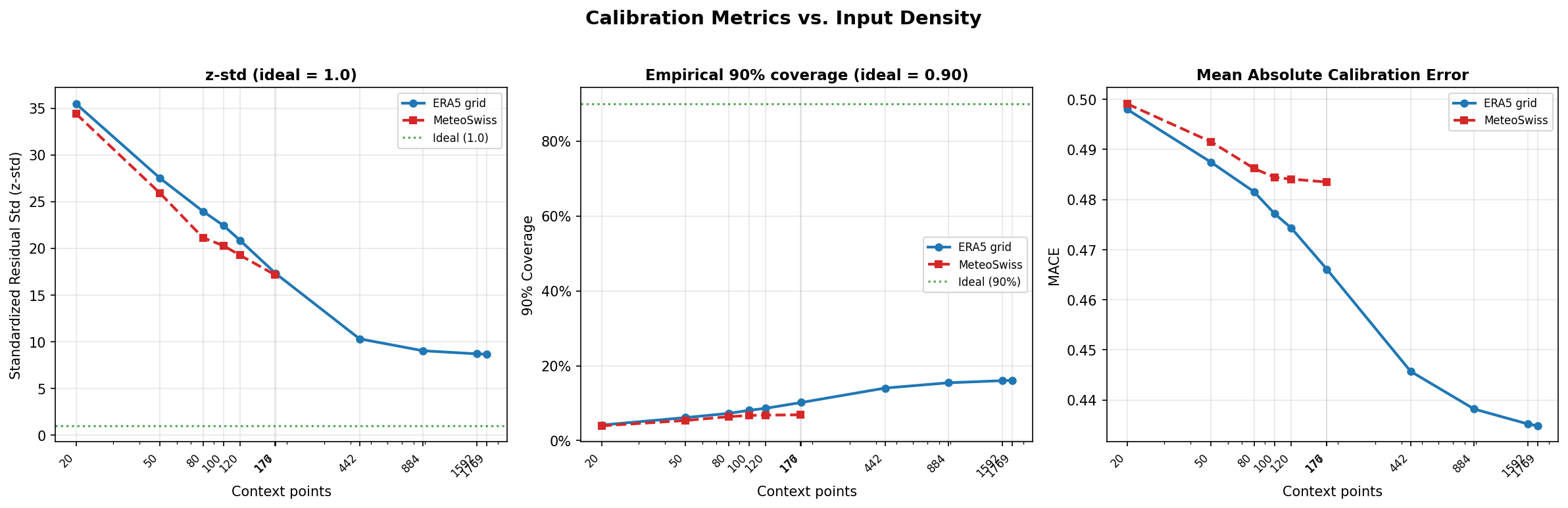}
    \caption{Calibration metrics vs.\ input density. Left: standardised residual standard deviation ($z$-std; ideal~=~1). Centre: empirical 90\% prediction interval coverage (ideal~=~90\%). Right: Mean Absolute Calibration Error (MACE). The model is severely overconfident at all densities; overconfidence worsens as context becomes sparser.}
    \label{fig:calibration_vs_density}
\end{figure}

Figure~\ref{fig:calibration_vs_density} confirms that the model's overconfidence, already identified at full coverage ($z$-std $\approx$ 8.7), worsens dramatically as input density decreases: $z$-std reaches 35 at 20~ERA5 points and 34 at 20~MeteoSwiss stations. The 90\% prediction interval coverage, already far below the nominal 90\% at full coverage (16.5\%), drops to 4.3\% at 20~ERA5 points. This reinforces that the Gaussian likelihood training objective does not produce well-calibrated uncertainties, and the miscalibration is amplified when the model operates outside its training distribution.

Representative single-day prediction maps for all sparsity configurations---showing the progressive visual degradation of the input field, the model prediction, and the residual bias---are collected in Appendix~\ref{sec:sparse_predictions}.

\section{Discussion of Results}

This section revisits the four research questions posed in the introduction in light of the experimental evidence presented above.

\paragraph{Model viability.}
The ConvCNP, adapted from the UK-focused implementation of \citet{vaughan2022} and equipped with high-resolution Swiss topographic features, produces accurate downscaled temperature fields over Switzerland. The best model achieves an MAE of 1.31\,\textdegree C and an RMSE of 1.71\,\textdegree C across roughly 3\,650 holdout days, with near-zero mean bias ($-$0.05\,\textdegree C). The spatial structure of the predictions closely matches the MeteoSwiss ground truth, resolving Alpine valleys, ridges, and plateaus that are invisible at the ERA5-Land resolution. These results confirm that the architecture transfers successfully to the topographically complex Swiss domain.

\paragraph{Skill over interpolation.}
All primary models outperform bilinear interpolation of the ERA5-Land grid. Even the weakest configuration (single year, 10~epochs) achieves a CRPS-based skill of 0.245, while the ten-year model reaches 0.524---reducing the expected probabilistic error by more than half relative to the interpolation baseline. This demonstrates that the ConvCNP learns meaningful fine-scale structure that pure spatial interpolation cannot capture.

\paragraph{Feature importance.}
The ablation study reveals a clear hierarchy of feature contributions. The elevation MLP and its topographic inputs are \emph{indispensable}: removing them causes the model to diverge catastrophically (MAE~$\approx$~802\,\textdegree C), confirming that the MLP is not a refinement but the essential component bridging the coarse-to-fine resolution gap. The Topographic Position Index provides a small but measurable benefit (MAE increases by 0.02\,\textdegree C without it), consistent with its role as a secondary terrain descriptor. The seasonal features present a more nuanced picture: removing them from the MLP slightly \emph{improves} performance (MAE 1.29 vs.\ 1.31\,\textdegree C), suggesting a mild regularisation benefit from reducing the MLP's input dimensionality. Furthermore, the two seasonal ablations---removing seasonal features everywhere versus only from the MLP---produce bit-identical results, indicating that the CNN entirely ignores the explicit seasonal channels in its input. The temperature field itself evidently carries sufficient seasonal signal for the CNN, while the MLP benefits from a leaner feature set.

\paragraph{Scaling behaviour.}
Two scaling axes were investigated: training duration and data volume. Increasing epochs from 10 to 30 yields a 22\% MAE reduction (2.09 to 1.64\,\textdegree C), but extending to 100~epochs provides only marginal further gains (1.62\,\textdegree C), indicating that the single-year model saturates its representational capacity within roughly 30~epochs. Importantly, no overfitting is observed even at 100~epochs. In contrast, moving from one year to ten years of training data at the same epoch count produces the largest single improvement (MAE from 1.64 to 1.31\,\textdegree C; Skill from 0.410 to 0.524), demonstrating that climatological diversity---exposure to varied synoptic situations and seasonal extremes---is more valuable than additional gradient steps on limited data. This suggests that further performance gains may be achievable by training on the full ERA5-Land archive (1950--present).

\paragraph{Robustness to sparse input.}
The model degrades gracefully when the ERA5-Land input grid is subsampled: at 50\% density the MAE increases by only 3.8\%, and positive skill is maintained down to approximately 10\% of the grid (${\sim}$177~points). Below this threshold, skill turns negative and the model performs worse than bilinear interpolation. Replacing the ERA5 grid entirely with MeteoSwiss station observations (off-grid, snapped to the nearest grid cell) yields negative skill at all tested densities, accompanied by a persistent warm bias of ${\sim}$2\,\textdegree C. This reflects both the distribution shift (the model was trained on dense gridded input) and the valley-biased sampling of the Swiss station network. The uncertainty estimates, which are highly informative at full coverage ($\sigma$--error correlation of 0.74), collapse to near-zero correlation below 10\% density and become anti-correlated for off-grid input.

\paragraph{Uncertainty calibration.}
Despite the model's strong point-prediction performance, its uncertainty estimates are severely overconfident across all configurations: the predicted $\sigma$ underestimates the true error magnitude by roughly an order of magnitude ($z$-std $\approx$ 8.7 at full coverage). The 90\% prediction interval covers only 16.5\% of observations. This miscalibration, consistent across training configurations and worsening with sparser input, is a known limitation of the Gaussian likelihood objective, which optimises overall fit rather than interval coverage. Nonetheless, the \emph{ranking} of uncertainties remains informative: the model reliably assigns higher $\sigma$ to locations where it makes larger errors, even though the absolute magnitudes are too small.
\chapter{Conclusion and Future Work}

\section{Conclusion}

This project set out to assess whether Convolutional Conditional Neural Processes---originally demonstrated for UK climate downscaling by \citet{vaughan2022}---can produce accurate, uncertainty-aware temperature predictions over the topographically complex terrain of Switzerland. The answer is affirmative: the ConvCNP, equipped with a high-resolution elevation MLP, achieves an MAE of 1.31\,\textdegree C and a CRPS-based skill of 0.524, reducing the expected prediction error by more than half compared to bilinear interpolation of the ERA5-Land grid. The model resolves fine-scale spatial structure---Alpine valleys, ridges, and plateaus---that is invisible at the ${\sim}$11\,km input resolution.

Scaling experiments reveal that training data volume is the dominant lever for improvement: moving from one to ten years of data yields a larger gain than tripling the number of training epochs, and no overfitting is observed even at 100~epochs. The ablation study confirms that the elevation MLP is the essential architectural component---without it, the model diverges entirely---while TPI provides a small secondary benefit and explicit seasonal features in the MLP are slightly detrimental, suggesting a regularisation advantage from a leaner feature set.

The model degrades gracefully under sparse on-grid input, maintaining positive skill down to approximately 10\% of the ERA5-Land grid. However, zero-shot deployment on off-grid MeteoSwiss station observations does not achieve positive skill at any tested density, highlighting the distribution shift between the dense gridded training regime and sparse irregular inference. Uncertainty calibration remains an open challenge: all configurations are severely overconfident, though the ranking of predicted uncertainties is informative.

\section{Limitations}

\begin{enumerate}
    \item \textbf{Data breadth.} Compared to \citet{vaughan2022}, this project uses fewer input channels: only temperature, coordinates, and seasonal encodings for the CNN, plus topographic features for the MLP. The original study additionally included geopotential height, wind fields, and humidity as CNN input channels. The narrower feature set likely limits the model's ability to capture atmosphere-driven local effects such as F\"ohn winds.

    \item \textbf{Uncertainty calibration.} The model's predicted uncertainties are overconfident by roughly an order of magnitude ($z$-std $\approx$ 8.7; 90\% interval coverage of only 16.5\%). This is a known limitation (\citet{guo2017}) of the Gaussian log-likelihood training objective, which concentrates gradient signal around the mean rather than penalising miscalibrated quantiles uniformly. Increasing training data volume from one to ten years did not resolve this issue, suggesting it is structural rather than data-limited.

    \item \textbf{Evaluation against a gridded product.} The MeteoSwiss ground truth is itself an interpolated product, not raw station measurements, and it does not align to the training grid. Errors in the ground-truth interpolation propagate into the evaluation metrics. And the bias of using a portion of the ground-truth to predict all the ground truth also seeps in.

    \item \textbf{Elevation-agnostic CNN.} As mentioned in the background section, in this project all topographic information is supplied exclusively through the elevation MLP, not the CNN itself. elevation-agnostic. This architectural difference may partly explain why the elevation MLP is so critical in our configuration: it bears the full burden of topographic correction, with no elevation-aware spatial context from the CNN to build upon.

    \item \textbf{Single variable.} Only daily maximum temperature was downscaled. For scoping reasons, minimum and average were not explored. The architecture supports precipitation via a Bernoulli--Gamma output layer, but this path was not exercised and may require additional work to wire correctly.
\end{enumerate}

\section{Future Work}

The findings and limitations of this project suggest several directions for further development.

\paragraph{Geopotential as a CNN channel.}
The most immediate architectural change would be to include the ERA5-Land geopotential field directly as a CNN input channel, following \citet{vaughan2022}. This would give the convolutional encoder access to coarse elevation context during spatial processing, potentially reducing the burden on the elevation MLP and improving predictions in areas where the coarse grid poorly represents the terrain.

\paragraph{Improved uncertainty calibration.}
Replacing the Gaussian log-likelihood with a CRPS-based loss function could address the overconfidence at its root: CRPS penalises miscalibrated quantiles more evenly across the distribution, tending to produce wider, better-calibrated prediction intervals.

\paragraph{Additional input channels.}
The model currently lacks atmospheric context beyond temperature. Integrating wind fields, humidity, and surface pressure as CNN channels---as well as terrain-derived features such as aspect and land cover---would provide the model with physically relevant information about local meteorological processes. These additions follow the extension points documented in Appendix~\ref{sec:adding-features}.

\paragraph{Multi-scale topographic features.}
Currently only TPI at 500\,m radius is used. Including TPI at additional scales (e.g.\ 2\,000\,m, also provided in swisstopo DHM25) would encode both local and wider topographic context, allowing the model to distinguish situations such as a hill within a deep valley from a hill on open plains---a distinction relevant for phenomena like cold-air pooling and valley inversions.

\paragraph{Larger elevation MLP.}
The observation that removing seasonal features from the MLP slightly improves performance suggests the MLP may be capacity-constrained. Increasing its width or depth could allow it to capture more complex nonlinear interactions between elevation, terrain shape, and atmospheric conditions, and might also improve uncertainty calibration by giving the network more expressive power for the $\sigma$ output.

\paragraph{Temporal context windows.}
The current model processes each day independently. Providing a temporal window of consecutive grids as input would allow the CNN to track the movement and evolution of weather systems---e.g.\ the progression of cold fronts across the Alps---potentially improving predictions during rapidly changing conditions.

\paragraph{Climate-aware temporal encoding.}
Including a year embedding alongside the seasonal encoding would allow the model, when trained over multi-decadal archives, to account for long-term climate trends such as rising temperatures and glacier retreat---effects that are relevant over the full ERA5-Land record (1960--present) but invisible to a purely seasonal encoding.

\paragraph{Hourly resolution and diurnal cycles.}
Switching from daily to hourly datasets and including time-of-day embeddings would be necessary if this technology is to support real-time prediction from live station feeds---a direction of potential interest to SDSC and operational meteorological services.

\paragraph{Precipitation downscaling.}
The ConvCNP architecture supports precipitation via a Bernoulli--Gamma output layer, as demonstrated by \citet{vaughan2022}. Extending the pipeline to precipitation downscaling is a natural next step, though it introduces additional complexity from the mixed discrete-continuous nature of rainfall.

\paragraph{Native off-grid support.}
Deploying the model operationally on sparse station data requires moving beyond the current snap-to-grid approach. Architectural options---including a true set convolution encoder, attention-based mechanisms, and dual-stream architectures---are discussed in detail in Appendix~\ref{sec:out-of-grid}.

\paragraph{Training infrastructure.}
Renku GPU sessions are occasionally interrupted during multi-day training runs for inscrutable reasons. Implementing checkpoint-based fold resumption---allowing training to restart from the middle of a fold rather than from the beginning---would improve robustness and reduce wasted compute.

\renewcommand{\bibname}{References}

\nocite{*}
\bibliographystyle{unsrtnat}
\bibliography{references}
\addcontentsline{toc}{chapter}{References}

\appendix
\chapter{Appendix}
\section{Code Structure and Responsibilities}
\label{sec:code-structure}

All the code produced, models trained, configurations used, plots and statistics generated can be found at \url{https://github.com/franciscopassos/convNPClimate/}.

\subsection{Repository Layout}

The repository is organized as follows:

\dirtree{%
  .1 convNPClimate/.
  .2 convCNP/\DTcomment{Original Vaughan et al.\ model code}.
  .3 models/\DTcomment{Neural network architectures}.
  .3 training/\DTcomment{Training and loss functions}.
  .3 validation/\DTcomment{Fold splitting and distance utilities}.
  .2 params.py\DTcomment{Configuration and hyperparameters}.
  .2 datasets.py\DTcomment{Data loading and preprocessing}.
  .2 model\_factory.py\DTcomment{Model construction}.
  .2 inference.py\DTcomment{Prediction utilities}.
  .2 metrics.py\DTcomment{Evaluation metrics (CRPS, skill)}.
  .2 visualization.py\DTcomment{Plotting and calibration diagrams}.
  .2 cloud\_sync.py\DTcomment{Renku cloud storage sync}.
  .2 json\_utils.py\DTcomment{Dataclass serialization}.
  .2 training\_notebook.ipynb\DTcomment{Training pipeline}.
  .2 predictions.ipynb\DTcomment{Inference and evaluation}.
  .2 trained\_models/\DTcomment{Saved checkpoints}.
  .2 datasets/\DTcomment{ERA5, MeteoSwiss, topography data}.
  .2 report-src/\DTcomment{This report}.
}

\subsection{Original Vaughan Code (\texttt{convCNP/})}

The \texttt{convCNP/} directory contains the core ConvCNP implementation from \citet{vaughan2022}, with minimal modifications on our part (highlighted below, and all diffs in repo at \url{https://github.com/franciscopassos/convNPClimate/blob/master/convcnp-diffs.txt}).
Its internal structure is as follows:

\begin{description}

  \item[\texttt{models/encoder.py}] The \emph{set convolution} encoder. Maps context observations onto a regular grid and produces a density channel indicating where data is present.

  \item[\texttt{models/cnn.py}] The convolutional decoder: a residual CNN that transforms the gridded encoder output into a spatial feature map.

  \item[\texttt{models/mlp.py}] A multi-layer perceptron block used as an intermediate mapping between the CNN output and the final distribution layer.

  \item[\texttt{models/final\_layers.py}] Distribution-specific output layers.
  \texttt{GaussianFinalLayer} outputs mean and standard deviation for temperature predictions; \newline \texttt{GammaFinalLayer} outputs the parameters of a Bernoulli--Gamma mixture for precipitation.
  Both use an RBF kernel interpolation (\texttt{ParamLayer}) to map from the regular grid back to arbitrary target point locations.

  \item[\texttt{models/models.py}] The base ConvCNP model classes: \texttt{TmaxBias\-ConvCNP} for temperature (Gaussian output) and \texttt{GammaBias\-ConvCNP} for precipitation (Bernoulli--Gamma output).
  Each composes the encoder, CNN decoder, MLP, and final layer into a full forward pass.

  \item[\texttt{models/elev\_models.py}] Elevation-enhanced variants: \texttt{Tmax\-Bias\-ConvCNP\-Elev} and \texttt{Gamma\-Bias\-ConvCNP\-Elev}.
  These append an elevation MLP that refines the base model's predictions using topographic features (true elevation, elevation difference from ERA5, mTPI) and, optionally, seasonal features ($\cos$ and $\sin$ of day-of-year).
  \begin{itemize}
      \item[] \textbf{Changes introduced in this project:}
      \begin{itemize}
          \item Wired new hyperparameters for ablation: \texttt{use\_seasonal\_in\_mlp}
      \end{itemize}
  \end{itemize} 

  \item[\texttt{training/training\_elev.py}] The main training loop supporting $k$-fold cross-validation with early stopping. Computes per-fold evaluation metrics (MAE, Pearson and Spearman correlation, NLL) and saves the best checkpoint per fold. \nopagebreak
    \begin{itemize} \nopagebreak
      \item[] \textbf{Changes introduced in this project:}
      \begin{itemize}
          \item Wired previously ignored hyperparameters: \texttt{device}, \texttt{batch\_size} and \texttt{patience}.
          \item Wired hyperparameter for ablation: \texttt{device}, \texttt{batch\_size} and \texttt{patience}.
          \item Wired \texttt{seasonal} features in for training.
          \item Replaced hardcoded fold indices with a calculation based on \texttt{fold} and \texttt{n\_folds}.
          \item Return more metrics (correlation scores).
          \item Append statistics to a CSV file once per epoch.
          \item Extracted logic into \texttt{get\_fold\_holdout\_indices()} and \texttt{select\_holdout\_day()} functions for reuse during prediction.
          \item Bug fix: replaced use of \texttt{np.float()} (deprecated) with \texttt{.float()}.
          \item Bug fix: suppressed warnings on empty slices.
      \end{itemize}
  \end{itemize} 

  \item[\texttt{training/loss\_functions.py}] Negative log-likelihood losses: Gaussian (\texttt{gll}) for temperature and Bernoulli--Gamma (\texttt{gamma\_ll}) for precipitation.
  Both handle missing target values (NaN masking).

  \item[\texttt{models/utils.py}] Utility functions shared across model components: context mask generation, squared-distance computation between target points and grid cells, and a numerically stable softplus (\texttt{force\_positive}) used to enforce positive distribution parameters.
  \begin{itemize}
      \item[] \textbf{Changes introduced in this project:}
      \begin{itemize}
          \item Bug fix: hardcoded cuda() calls. Now using a hyperparameter.
      \end{itemize}
  \end{itemize} 

\item[\texttt{training/utils.py}] Data preparation helpers for the training loop: epoch-level shuffling, fold splitting into train/held-out batches, value extraction functions (\texttt{get\_value\_tmax}, \texttt{get\_sigma\_tmax}), and the precipitation rain mask (\texttt{make\_r\_mask}).
  \begin{itemize}
      \item[] \textbf{Changes introduced in this project:}
      \begin{itemize}
            \item Wired previously ignored hyperparameters: \texttt{batch\_size}, \texttt{device}.
            \item Wired hyperparameter: \texttt{seasonal}.
            \item Bug fix: coercing a type to int, to prevent accidentally generating float indices which would cause a crash.
            \item Exposed uncertainty via a function \texttt{get\_sigma\_tmax()}.
      \end{itemize}
  \end{itemize} 

\item[\texttt{validation/utils.py}] Model loading and inference helpers for validation: reconstructs a model from a checkpoint, generates predictions on held-out data, and computes target-to-grid distances.
  \begin{itemize}
      \item[] \textbf{Changes introduced in this project:}
      \begin{itemize}
            \item Wired previously ignored hyperparameters: \texttt{in\_channels}, \texttt{n\_blocks}, \texttt{kernel\_size} and \texttt{length\_scale} (these were hardcoded).
      \end{itemize}
  \end{itemize} 

\end{description}

\subsection{Python Modules Added by This Project}

The following modules, located at the repository root, were developed as part of this capstone project to provide the scaffolding needed for systematic experimentation, evaluation, and reproducibility.

\begin{description}

  \item[\texttt{params.py}]
  Defines the \texttt{Params} dataclass, which centralizes all configurable hyperparameters and paths: model architecture settings (channels, blocks, kernel size), training settings (epochs, batch size, learning rate, patience, folds), ablation flags for seasonal features, elevation and mTPI, and output paths.
  It also provides environment-detection utilities (\texttt{is\_renku()}, \texttt{configure\_renku\_cuda()}, \texttt{select\_device()}) and a reproducibility helper (\texttt{set\_seed()}).
  Params instances are serializable to JSON via \texttt{save\_json()} and \texttt{load\_json()}, ensuring that every trained model has a complete record of its configuration.

  \item[\texttt{datasets.py}]
  This module is responsible for all data ingestion and preprocessing.
  Key capabilities include:
  \begin{itemize}
    \item Loading ERA5-Land NetCDF files into normalized context tensors, with configurable auxiliary channels (latitude, longitude, seasonal encodings);
    \item Coordinate conversion between WGS84 and Swiss LV95 projections;
    \item Loading high-resolution topography (DEM and mTPI) from Zarr format;
    \item Preparing MeteoSwiss ground truth as target tensors, including interpolation of high-resolution elevation features to target locations;
    \item Building sparse context representations for limited input experiments;
    \item Computing bilinear ERA5 interpolations to station locations as a deterministic baseline.
  \end{itemize}

  \item[\texttt{model\_factory.py}]
  A factory function \texttt{build\_model()} that constructs the full model stack from a \texttt{Params} object. It returns a \texttt{(model, loss\_fn, get\_value\_fn)} tuple, abstracting the differences between temperature and precipitation model variants. Also provides \texttt{load\_model\_checkpoint()} for restoring trained models. This module allows consistent use across training and prediction. At training time we persist the metadata.json hyperparameters, and at inference time we use it to recreate and load the model. 

  \item[\texttt{inference.py}]
  Provides \texttt{predict\_single\_day()} for generating and denormalizing predictions for a single day and \texttt{predict\_holdout\_fold()} for doing the same across all holdout days in a cross-validation fold.

  \item[\texttt{metrics.py}]
  Computes per-pixel evaluation grids: MAE, RMSE, bias, CRPS , and a skill score relative to the deterministic ERA5 baseline.
  Results are collected in a \texttt{PerPixelMetrics} dataclass.

  \item[\texttt{visualization.py}]
  A comprehensive plotting library ($\approx$1\,100 lines) covering:
  \begin{itemize}
    \item Single-day prediction comparisons (ERA5 input, ground truth, model output, residuals);
    \item Per-pixel error maps (MAE, RMSE, bias) with spatial rendering;
    \item Uncertainty analysis (predicted $\sigma$ vs.\ observed error);
    \item Probabilistic calibration: Q--Q plots using the Probability Integral Transform (PIT) and reliability diagrams with MACE and RMSCE metrics;
    \item CRPS and skill-score maps;
    \item Training-curve visualization across folds.
  \end{itemize}

  \item[\texttt{cloud\_sync.py}]
  Handles synchronization of datasets and trained models between local storage and the Renku network mount, enabling seamless transitions between local development and cloud training. (Note: using the remote mount directly from Polybox was timing out.)

  \item[\texttt{json\_utils.py}]
  Serialization helpers for dataclasses containing NumPy arrays and PyTorch device objects, used to persist \texttt{Params} and \texttt{Era5Metadata} as JSON.

\end{description}

\subsection{Main Notebooks}

The two primary notebooks serve as the entry points for training and evaluation, respectively.

\subsubsection{Training (\texttt{training\_notebook.ipynb})}

This notebook orchestrates the end-to-end training pipeline:

\begin{enumerate}
  \item \textbf{Configuration.} A \texttt{Params} object is instantiated with the desired hyperparameters and ablation flags.
  The notebook auto-detects the execution environment (local vs.\ Renku) and selects the appropriate compute device.

  \item \textbf{Data loading.} ERA5-Land temperature and geopotential fields are loaded and normalized.
  Seasonal features (cosine and sine of day-of-year) are optionally appended as additional input channels.
  High-resolution DEM and mTPI grids are loaded from Zarr.
  MeteoSwiss station observations are prepared as target tensors with associated elevation features.

  \item \textbf{Training loop.} For each of the $k$ cross-validation folds, a fresh model is built via \texttt{model\_factory}, random seeds are reset for reproducibility, and \texttt{train\_elev()} is called.
  Per-fold statistics (loss, MAE, correlations) are logged to CSV and, on Renku, checkpoints are synced to cloud storage after each fold completes.

  \item \textbf{Summary.} Per-fold statistics are merged into a single CSV and training curves are visualized.
\end{enumerate}

The notebook saves all outputs---checkpoints, \texttt{params.json}, \texttt{metadata.json}, and statistics---to a timestamped directory under \texttt{trained\_models/}, ensuring full reproducibility of every experiment.

\subsubsection{Prediction and Evaluation (\texttt{predictions.ipynb})}

This notebook loads a trained model and performs comprehensive evaluation:

\begin{enumerate}
  \item \textbf{Configuration.} The user specifies a trained model by name and selects the input mode: \texttt{ERA5\_GRID} (full ERA5 context, optionally sparsified) or \texttt{METEOSWISS} (station-only context placed onto an otherwise empty grid).

  \item \textbf{Data reconstruction.} The training configuration (\texttt{Params} and \texttt{Era5Metadata}) is loaded from the saved JSON files to ensure identical preprocessing. All datasets are reloaded accordingly.

  \item \textbf{Per-fold inference.} For each cross-validation fold, the best checkpoint is loaded and predictions are generated for all holdout days. A representative day is selected for single-day visualizations (ERA5 input, ground truth, prediction, residuals).

  \item \textbf{Combined evaluation.} Predictions from all folds are aggregated to compute per-pixel metrics: MAE, RMSE, bias, CRPS, and skill score relative to the ERA5 baseline.
  Correlations with altitude and mTPI are analysed.

  \item \textbf{Calibration analysis.} Uncertainty calibration is assessed via Q--Q plots (based on the PIT) and reliability diagrams that compare predicted confidence levels against observed coverage.

  \item \textbf{Export.} All metrics are saved to a JSON file for downstream comparison across experiments. All plots are saved to a plots subfolder in the \texttt{trained\_models/<model\_name>} directory.
\end{enumerate}

\section{Dataset Expectations}
\label{sec:dataset-expectations}

The pipeline expects three dataset families stored under a \texttt{datasets/} directory.
All gridded files are NetCDF; topography is stored as Zarr.

\dirtree{%
  .1 datasets/.
  .2 ERA5\_Land/.
  .3 temperature/ (unused)\DTcomment{\texttt{t2m-\{1960..2023\}.nc}}.
  .3 max\_temperature/\DTcomment{\texttt{t2m\_max-\{1960..2023\}.nc}}.
  .3 min\_temperature/ (unused)\DTcomment{\texttt{t2m\_min-\{1960..2023\}.nc}}.
  .3 precipitation/ (unused)\DTcomment{\texttt{tp-\{1960..2023\}.nc}}.
  .3 geopotential/\DTcomment{\texttt{era5\_land\_geopotential.nc}}.
  .2 MeteoSwiss/.
  .3 TabsD\_v2.0\_swiss.lv95/ (unused)\DTcomment{Daily mean temperature}.
  .3 TmaxD\_v2.0\_swiss.lv95/\DTcomment{Daily max temperature}.
  .3 TminD\_v2.0\_swiss.lv95/ (unused)\DTcomment{Daily min temperature}.
  .3 RhiresD\_v2.0\_swiss.lv95/ (unused)\DTcomment{Daily precipitation}.
  .2 topo\_subset.zarr/\DTcomment{swisstopo DHM25 derivatives}.
}

The folders marked unused were not used in the course of this specific project. All temperature ones can be easily switched in. The precipitation one was not used and because it exposes two distribution parameters, might require additional work to make sure it all is wired through correctly.

\begin{description}
  \item[ERA5-Land] Reanalysis fields from ECMWF, one NetCDF file per year.
  The temperature and geopotential files are required; the others are optional depending on the variable being downscaled.
  Source: \url{https://confluence.ecmwf.int/display/CKB/ERA5-Land:+data+documentation}.

  \item[MeteoSwiss] Gridded ground-truth products on the Swiss LV95 coordinate system, one file per year.
  Each subfolder contains daily grids at $\approx$1\,km resolution.
  Source: \url{https://www.meteoswiss.admin.ch}.

  \item[Topography (\texttt{topo\_subset.zarr})] A Zarr store derived from the swisstopo DHM25 digital elevation model.
  Contains the DEM, coordinates, Topographic Position Index at 500\,m and 2000\,m scales (TPI\_500M, TPI\_2000M), slope derivatives, and valley-norm features.
\end{description}

Loading is handled by the \texttt{datasets.py} module (Section~\ref{sec:code-structure}), which normalizes the ERA5 fields, converts coordinates between WGS84 and LV95, and interpolates topographic features to station locations. The raw files can also be opened directly with \texttt{xarray} (disrecommended):

\begin{verbatim}
import xarray as xr
era5 = xr.open_dataset("datasets/ERA5_Land/temperature/t2m-1960.nc")
mch  = xr.open_dataset("datasets/MeteoSwiss/TabsD_v2.0_swiss.lv95/"
         "TabsD_ch01r.swiss.lv95_196101010000_196112310000.nc")
topo = xr.open_zarr("datasets/topo_subset.zarr")
\end{verbatim}

\section{Running on Renku}
\label{sec:running-on-renku}

The training pipeline can run on the Renku cloud platform to take advantage of GPU resources.
Two setup steps are required: mounting external storage and configuring CUDA.

\subsection{Mounting Datasets and Model Storage via Polybox}

Renku sessions have limited local disk space, so datasets and trained models are stored on ETH Polybox and mounted into the session.

\begin{description}
  \item[Datasets (read-only).] Place all dataset files (Section~\ref{sec:dataset-expectations}) in a Polybox folder and mount it into the Renku session at a mount-point called \texttt{datasets-mount}, using your ETH credentials.
  On first use, the training notebook calls \newline \texttt{sync\_from\_cloud\_if\_needed()}, which copies the data from the network mount to the local \texttt{datasets/} directory.
  Subsequent runs skip files that are already present locally.
  This local-copy step was necessary because reading NetCDF files directly from the remote mount caused frequent I/O timeouts.

  \item[Trained models (read-write).] Create a second, initially empty Polybox folder and mount it read-write at \texttt{trained-models-mount}.
  After each cross-validation fold completes, the training notebook calls \texttt{sync\_to\_cloud\_if\_needed()}, which copies checkpoints, parameters, and statistics to the remote mount.
  This ensures that results are persisted even if the Renku session is terminated.
\end{description}

When the \texttt{Params.RUN\_TYPE} is set to \texttt{'cloud'} (detected automatically via \texttt{is\_renku()}), the sync functions are activated; when set to \texttt{'local'}, they are no-ops.

\subsection{CUDA Configuration for MIG GPUs}

Renku GPU sessions use NVIDIA Multi-Instance GPU (MIG) partitions.
PyTorch's CUDA caching allocator has an ``expandable segments'' feature that relies on NVML calls which fail on MIG devices, causing crashes at the first tensor allocation.

To work around this, \texttt{configure\_renku\_cuda()} must be called \emph{before} \texttt{import torch}.
It sets the environment variable:

\begin{verbatim}
PYTORCH_CUDA_ALLOC_CONF=expandable_segments:False,max_split_size_mb:512
\end{verbatim}

This disables the incompatible allocator feature and caps the maximum split size to avoid memory fragmentation.
Both training and prediction notebooks call this function in their first cell, before any other PyTorch import.

\section{Training and Predicting}
\label{sec:training-predicting}

\subsection{Training}

\subsubsection{Configuration}

All training parameters are set in the second cell of \texttt{training\_notebook.ipynb} by instantiating a \texttt{Params} object.
The key fields are:

\begin{itemize}
  \item \texttt{VARIABLE} --- the target variable (\texttt{'tmax'} or \texttt{'precip'}).
  \item \texttt{DATA\_YEAR\_START} --- first year of training data (\texttt{None} for the full archive, e.g.\ \texttt{2023} for a single-year run).
  \item \texttt{TRIAL\_NAME} --- a human-readable name that becomes the output subdirectory.
  \item \texttt{N\_EPOCHS}, \texttt{BATCH\_SIZE}, \texttt{LR}, \texttt{PATIENCE}, \texttt{N\_FOLDS} --- standard training hyperparameters.
  \item Ablation flags: \texttt{SEASONAL\_FEATURES}, \texttt{SEASONAL\_FEATURES\_IN\_MLP}, \texttt{USE\_ELEVATION}, \texttt{USE\_MTPI}.
\end{itemize}

\texttt{RUN\_TYPE} and \texttt{DEVICE} are auto-detected and should not normally be changed.
After data loading, the notebook calls \texttt{PARAMS.with\_in\_channels()} to set the number of CNN input channels automatically.

\subsubsection{Running}

Execute all cells in order.
The notebook loads the datasets, builds a fresh model per fold via \texttt{model\_factory.build\_model()}, and calls \texttt{train\_elev()} for each fold.
On Renku, checkpoints are synced to the remote mount after every fold (see Section~\ref{sec:running-on-renku}).

\subsubsection{Training Outputs}

All outputs are written to \texttt{trained\_models/<TRIAL\_NAME>/}:

\begin{description}
  \item[\texttt{params.json}] The full \texttt{Params} object, including data paths. Reloaded by the prediction notebook to ensure identical configuration.

  \item[\texttt{metadata.json}] ERA5 normalization statistics (global mean and standard deviation, latitude/longitude bounds, coordinate arrays). Required to denormalize predictions back to physical units.

  \item[\texttt{model\_fold\_\{0..N-1\}}] PyTorch checkpoint for the best epoch of each fold (selected by validation NLL). Contains \texttt{model\_state\_dict}, optimizer state, and epoch number.

  \item[\texttt{stats\_fold\{0..N-1\}.csv}] Per-epoch training statistics for each fold: MAE, Pearson and Spearman correlations, training and validation NLL.

  \item[\texttt{stats.csv}] All per-fold CSVs concatenated into a single file for convenience.

  \item[\texttt{trainingstats\_*.png}] Training curve plots (per fold and combined average).
\end{description}

\subsection{Prediction and Evaluation}

\subsubsection{Configuration}

The prediction notebook (\texttt{predictions.ipynb}) has three configuration variables set near the top:

\begin{itemize}
  \item \texttt{MODEL\_NAME} --- must match a directory under \texttt{trained\_models/}.
  \item \texttt{INPUT\_DATA} --- either \texttt{'ERA5\_GRID'} (full or sparsified ERA5 context) or \texttt{'METEOSWISS'} (station observations placed onto the ERA5 grid, with the rest set to NaN).
  \item \texttt{SPARSITY\_TYPE} and its value --- controls how much context data is retained: \texttt{'PERCENT'} with a fraction (e.g.\ \texttt{0.5} keeps 50\% of grid cells) or \texttt{'ABSOLUTE'} with a fixed count (e.g.\ \texttt{100} keeps 100 cells). Setting \texttt{SPARSITY\_PERCENT=1.0} uses the full grid.
\end{itemize}

All other settings (\texttt{Params}, normalization statistics, architecture) are loaded from the model's saved \texttt{params.json} and \texttt{metadata.json}.

\subsubsection{Running}

Execute all cells in order.
The notebook reloads the datasets using the saved configuration, builds the context tensor according to the chosen input mode and sparsity, and then for each fold:

\begin{enumerate}
  \item Loads the best checkpoint for that fold.
  \item Generates predictions for all holdout days.
  \item Plots a single representative holdout day (ERA5 input, ground truth, prediction, residuals).
\end{enumerate}

After all folds, holdout predictions are concatenated (the folds cover non-overlapping time periods) and the combined evaluation is computed: error maps, uncertainty calibration, CRPS, skill scores, and correlation analyses.

\subsubsection{Prediction Outputs}

Outputs are written to \texttt{trained\_models/<MODEL\_NAME>/}:

\begin{description}
  \item[\texttt{plots/}] All visualization PNGs, named with a prefix encoding the input configuration.
  The prefix follows the pattern \texttt{\{INPUT\_DATA\}\_\{SPARSITY\_TYPE\}\_\{value\}\_}, e.g.\ \texttt{ERA5\_GRID\_PERCENT\_1.0\_error\_maps.png}.
  This allows multiple prediction runs (different sparsity levels) to coexist in the same directory.

  The plots include:
  \begin{itemize}
    \item Per-fold single-day predictions (\texttt{fold\{N\}\_prediction.png});
    \item Error maps: MAE, RMSE, bias (\texttt{error\_maps.png});
    \item Uncertainty vs.\ error scatter (\texttt{uncertainty\_vs\_error.png});
    \item Altitude and mTPI correlation plots \newline (\texttt{altitude\_vs\_error.png}, \texttt{mtpi\_vs\_error.png});
    \item Temporal error analysis (\texttt{doy\_vs\_error.png});
    \item CRPS map (\texttt{crps\_map.png});
    \item Skill score map (\texttt{skill\_map.png});
    \item Q--Q calibration plot (\texttt{qq\_calibration.png});
    \item Reliability diagram (\texttt{reliability\_diagram.png}).
  \end{itemize}

  \item[\texttt{metrics-\{INPUT\}-\{SPARSITY\}-\{value\}.json}] A JSON file summarizing all scalar metrics: overall MAE, RMSE, bias, CRPS, skill score, correlation coefficients (uncertainty--error, altitude--error, mTPI--error, day-of-year--error), and calibration statistics (PIT mean/std, coverage at 50\%/90\%/95\%, MACE).
  One file is produced per prediction run, enabling programmatic comparison across experiments and sparsity levels.
\end{description}

\section{How to Add New Features to the Model}
\label{sec:adding-features}

The model has three distinct extension points for new input features, each with a different data shape and code path.
This section walks through each one, using the seasonal and topographic features already in the codebase as concrete examples.

\subsection{Extension Point 1: CNN Input Channels}

CNN channels are \emph{gridded, per-timestep} features that share the ERA5 spatial grid (29$\times$61).
They flow through the set convolution encoder and the ResNet decoder, giving the CNN spatial context about each feature.
The seasonal cosine and sine encodings were added this way.

\paragraph{What to change.}

\begin{enumerate}
  \item \textbf{\texttt{datasets.py}, \texttt{load\_era5\_data()}.}
  Compute the new feature as a tensor of shape \texttt{(time, lat, lon)} and append it to \texttt{channel\_list} before the final concatenation.
  For seasonal features, this is where $\cos(2\pi d/365)$ and $\sin(2\pi d/365)$ are broadcast to the full grid and appended as channels~4 and~5:

\begin{verbatim}
  # datasets.py, inside load_era5_data()
  cos_time = np.cos(time_rads)
  sin_time = np.sin(time_rads)
  cos_channel, sin_channel, _ = xr.broadcast(
      cos_time, sin_time, norm_data)

  channel_list = [norm_data, lat_channel, lon_channel]
  if include_seasonal_embeddings:
      channel_list.extend([cos_channel, sin_channel])

  tensor_Z = xr.concat(channel_list, dim="channel")
\end{verbatim}

  \item \textbf{Everything else adapts automatically.}
  The training notebook calls \texttt{PARAMS.with\_in\_channels(tensor.shape[1])} after loading, so \texttt{IN\_CHANNELS} updates to reflect the new channel count.
  \texttt{model\_factory} passes it to the model constructor, and the encoder's linear layer (\texttt{nn.Linear(in\_channels*2, 128)}) adapts to the new width at initialization.
\end{enumerate}

No changes are needed in \texttt{params.py}, \texttt{model\_factory.py}, \texttt{encoder.py}, or \texttt{elev\_models.py}---the encoder is parameterized by \texttt{in\_channels} and adjusts automatically.

\subsection{Extension Point 2: Elevation MLP Features (Per-Location)}

Per-location features are \emph{static vectors attached to each target point}, such as the true elevation, the elevation difference from the ERA5 grid cell, and the Topographic Position Index.
They are concatenated with the CNN's base prediction ($\mu$, $\sigma$) and passed through the elevation MLP.
Because they are per-point rather than gridded, they can use data at much higher resolution than the ERA5 grid.

\paragraph{What to change (using mTPI as the example).}

\begin{enumerate}
  \item \textbf{\texttt{datasets.py}, \texttt{load\_high\_res\_topography()}.}
  Load the new raster (e.g.\ slope or aspect) from the Zarr store and return it alongside the DEM and TPI. (If this is an entirely separate feature, load if from the appropriate store separately.)

  \item \textbf{\texttt{datasets.py}, \texttt{prepare\_meteoswiss\_targets()}.}
  Interpolate the new raster to each target point's LV95 coordinates (bilinear), then append it to the feature stack.
  Currently the stack is \texttt{[true\_elev, elev\_diff, mTPI]}, giving a tensor of shape \texttt{(n\_points, 3)}.
  Adding a fourth feature changes this to \texttt{(n\_points, 4)}.
  Here is how mTPI was added:

\begin{verbatim}
  # datasets.py, inside prepare_meteoswiss_targets()
  if hi_res_tpi is not None:
      tpi = hi_res_tpi.interp(
          x=xr.DataArray(target_x_lv95, dims='point'),
          y=xr.DataArray(target_y_lv95, dims='point'),
          method='linear')
  else:
      tpi = xr.zeros_like(true_elev)

  tensor_e = xr.concat(
      [true_elev, elev_diff, tpi], dim="feature")
\end{verbatim}

  \item \textbf{\texttt{convCNP/models/elev\_models.py}.}
  Update the MLP input width.
  The elevation MLP input size is currently computed as:
  \[
    \underbrace{2}_{\mu,\,\sigma} + \underbrace{3}_{\text{elev features}} + \underbrace{2}_{\text{seasonal (if enabled)}} = 7
  \]
  Adding one elevation feature changes the 3 to a 4, so the MLP input width becomes 8 (or 6 without seasonal).

  This is set in the \texttt{\_\_init\_\_} of \texttt{TmaxBiasConvCNPElev} (and \texttt{GammaBiasConvCNPElev} for precipitation):

\begin{verbatim}
  # convCNP/models/elev_models.py, TmaxBiasConvCNPElev.__init__()
  # 2 (mu, sigma) + 3 (elev features) + 2 (seasonal) = 7
  elev_mlp_in_channels = 7 if use_seasonal_in_mlp else 5
  self.elev_mlp = MLP(elev_mlp_in_channels, 2,
      hidden_channels=64, hidden_layers=4)
\end{verbatim}

  No changes are needed in the \texttt{forward()} method---the concatenation already uses the full elevation tensor regardless of its width.

  \item \textbf{\texttt{params.py}.}
  Optionally add an ablation flag (e.g.\ \texttt{USE\_SLOPE} if you were adding slope derivative) so the feature can be toggled on and off for experiments.
  Pass it through to \newline \texttt{prepare\_meteoswiss\_targets()} and \texttt{elev\_models.py}.
\end{enumerate}

\subsection{Extension Point 3: Elevation MLP Features (Per-Timestep)}

Per-timestep features are \emph{scalars that vary by day but are the same for all target points}, such as the seasonal cosine and sine encodings.
At inference time they are broadcast (expanded) to every target point before concatenation with the base prediction and elevation features.

\paragraph{What to change (using seasonal features as the example).}

\begin{enumerate}
  \item \textbf{\texttt{datasets.py}, \texttt{compute\_seasonal\_features()}.}
  Compute the new feature as a tensor of shape \texttt{(n\_times, n\_features)}.
  For seasonal features (day of year, doy) this is \texttt{(n\_times, 2)} containing $[\cos(\text{doy}),\, \sin(\text{doy})]$:

\begin{verbatim}
  # datasets.py, compute_seasonal_features()
  times_pd = pd.to_datetime(time_coords)
  day_of_year = times_pd.dayofyear.values
  time_rads = (day_of_year - 1) / 365.0 * 2 * np.pi

  cos_doy = np.cos(time_rads)
  sin_doy = np.sin(time_rads)

  seasonal = np.stack([cos_doy, sin_doy], axis=1)  # (n_times, 2)
  return torch.tensor(seasonal, dtype=torch.float32, device=device)
\end{verbatim}

  \item \textbf{Training notebook.}
  Compute the feature after loading time coordinates and pass it as the \texttt{seasonal} argument to \texttt{train\_elev()}.
  If the feature should coexist with seasonal features, concatenate them into a single tensor before passing.

\begin{verbatim}
  # training_notebook.ipynb, after loading ERA5 data
  if PARAMS.SEASONAL_FEATURES_IN_MLP:
      seasonal_features = ds.compute_seasonal_features(
          time_coords, device=device)
  else:
      seasonal_features = None

  # Later, in the training loop:
  train_elev(..., seasonal=seasonal_features, ...)
\end{verbatim}

  \item \textbf{\texttt{convCNP/training/utils.py}, \texttt{get\_fold\_data()}.}
  Already handles splitting, shuffling, and batching the \texttt{seasonal} tensor by fold.
  No changes needed unless a separate parameter is introduced.

  \item \textbf{\texttt{convCNP/models/elev\_models.py}.}
  In \texttt{forward()}, per-timestep features are expanded from \texttt{(batch, n\_features)} to \texttt{(batch, n\_points, n\_features)} and concatenated along the feature dimension.
  Update the MLP input width as described above (the seasonal contribution changes from 2 to 2 + $n_{\text{new}}$).
  Here is how the broadcast and concatenation work in \texttt{forward()}:

\begin{verbatim}
  # convCNP/models/elev_models.py, TmaxBiasConvCNPElev.forward()
  elev_expanded = elev.repeat(batch_size, 1, 1)  # (batch, n_points, 3)

  if self.use_seasonal_in_mlp and seasonal is not None:
      # seasonal: (batch, 2) -> (batch, n_points, 2)
      seasonal_expanded = seasonal.unsqueeze(1).expand(
          -1, n_points, -1)
      out = torch.cat(
          [out, elev_expanded, seasonal_expanded], dim=2)
  else:
      out = torch.cat([out, elev_expanded], dim=2)

  out = self.elev_mlp(out)
\end{verbatim}
\end{enumerate}

\subsection{Summary}

Table~\ref{tab:extension-points} summarizes the three extension points and the files that require modification.

\begin{table}[ht]
\centering
\caption{Extension points for adding new features.}
\label{tab:extension-points}
\begin{tabular}{@{}llll@{}}
\toprule
\textbf{Extension point} & \textbf{Shape} & \textbf{Example} & \textbf{Files to modify} \\
\midrule
CNN channel        & (time, lat, lon)    & cos/sin of doy       & \texttt{datasets.py} only \\
MLP per-location   & (n\_points, $k$)    & elevation, mTPI      & \texttt{datasets.py}, \texttt{elev\_models.py} \\
MLP per-timestep   & (n\_times, $k$)     & seasonal encoding    & \texttt{datasets.py}, \texttt{elev\_models.py} \\
\bottomrule
\end{tabular}
\end{table}

\section{Considerations on supporting non-gridded station input}
\label{sec:out-of-grid}

\subsection{The Architectural Constraint}

The model's data flow is:

\begin{verbatim}
Gridded context (batch, channels, lat, lon)
  -> Encoder (Conv2d)            <- gridded assumption
  -> CNN decoder (Conv2d)        <- gridded assumption
  -> MLP (per grid-cell)
  -> RBF kernel (grid -> points) <- targets already off-grid
  -> Elevation MLP
  -> Point predictions
\end{verbatim}

The gridded assumption lives in the encoder and CNN decoder, which use \texttt{Conv2d}. The output side already handles off-the-grid targets via the RBF kernel. The ConvCNP paper \citep{vaughan2022} uses ERA5 gridded reanalysis as context, which is naturally gridded data. The architecture handles off-the-grid \emph{targets} natively, but the \emph{context} (input) side is where the constraint lies.

The question is therefore how to get non-gridded station data into the CNN pipeline.

\subsection{Current State: Snap-to-Grid for Prediction}

For prediction only (not training), a snap-to-grid approach is already implemented. The inference path works as follows:

\begin{enumerate}
    \item \texttt{find\_nearest\_era5\_indices} maps station coordinates to their nearest ERA5 grid cells.
    \item \texttt{build\_sparse\_context} places station values at those grid cells and fills unobserved cells with NaN.
    \item \texttt{predict\_single\_day} derives a binary mask from the NaN pattern (non-NaN $=$ observed, NaN $=$ unobserved) and runs the forward pass.
\end{enumerate}

This means a model trained on full ERA5 grids can be used at inference time with a sparse station-derived context, without any model changes.

\paragraph{Limitations of snap-to-grid.}
\begin{itemize}
    \item Loses sub-grid station positioning (stations are rounded to the nearest cell). In Switzerland, a single kilometre can correspond to a very high change in elevation, making this a severe limitation.
    \item Multiple stations falling in the same cell are averaged.
    \item With very sparse station networks, the $5 \times 5$ encoder kernel may not propagate information far enough (only ${\sim}2$ grid cells of reach).
    \item The model has only been trained on dense grids, so prediction quality with very sparse context suffers significantly.
\end{itemize}

\subsection{Options to Replace Snap-to-Grid}

The following options would require additional implementation work, either to improve input quality without model changes or to modify the model architecture itself.

\subsubsection{Option 1: True Set Convolution Encoder (ConvCNP-Native)}

The original ConvCNP formalism \citep{gordon2019} maps scattered points to a grid representation using a \textbf{set convolution}---essentially the reverse of what the final layer already does. This produces a gridded representation that the existing CNN decoder can consume without modification.

The current encoder does something conceptually similar (depthwise convolution with density normalization), but using \texttt{Conv2d}, which requires the input to already be on the grid. Replacing it with a kernel-based set convolution that takes station locations and values directly would architecturally clean. This would require rewriting the encoder.

\paragraph{Pros.}
\begin{itemize}
    \item Stays within the ConvCNP framework.
    \item Preserves exact station locations and handles arbitrary station counts.
    \item The rest of the pipeline (CNN, final layer, elevation MLP) remains unchanged.
    \item Translation equivariance is preserved.
\end{itemize}

\paragraph{Cons.}
\begin{itemize}
    \item Set convolution is $\mathcal{O}(N_{\text{stations}} \times N_{\text{grid}})$ per forward pass rather than the efficient \texttt{Conv2d}.
\end{itemize}

\subsubsection{Option 2: Attention-Based Encoder or Graph Neural Network}

Use AttentiveCNP or a graph neural network (GNN) where stations are nodes and edges connect nearby stations. This is a significant architectural change.

\paragraph{Pros.}
\begin{itemize}
    \item Natively handles irregular inputs.
    \item Attention can focus on the most relevant stations for each prediction.
\end{itemize}

\paragraph{Cons.}
\begin{itemize}
    \item Loses translation equivariance, a key property of the ConvCNP framework.
    \item Attention is $\mathcal{O}(n^2)$ in the number of stations---this will not scale if using massive grids as training input and may require smart approaches (e.g. subsampling).
\end{itemize}

\subsubsection{Option 3: Pre-Interpolate to a full higher-resolution grid}

Use an external spatial interpolation method (kriging, optimal interpolation, inverse distance weighting, thin-plate splines) to produce a complete gridded field from station data, then feed it into the model unchanged.

\paragraph{Pros.}
\begin{itemize}
    \item No model changes required.
    \item Produces a dense grid the CNN works with natively.
    \item Well-understood classical methods are available.
\end{itemize}

\paragraph{Cons.}
\begin{itemize}
    \item Interpolation error propagates into the model with no way to account for it.
    \item Uncertainty about data-sparse regions is lost.
\end{itemize}

\subsubsection{Option 4: Dual-Stream Architecture}

If some gridded auxiliary predictors are available (e.g.\ ERA5-Land, etc) alongside station observations, keep the existing CNN path for the gridded data but add a parallel point-data encoder (set convolution or attention-based) for the station values. Fuse the two representations on the grid before the decoder.

\paragraph{Pros.}
\begin{itemize}
    \item Leverages gridded reanalysis where available.
    \item Does not discard information from either source.
\end{itemize}

\paragraph{Cons.}
\begin{itemize}
    \item More complex architecture.
    \item Need to design the fusion strategy.
    \item Two encoder branches to train.
\end{itemize}

\section{Sparse Input Prediction Samples}
\label{sec:sparse_predictions}

The following figures show representative single-day predictions (fold~5) at selected sparsity levels, as evaluated in Section~\ref{sec:sparse}. Each row shows, from left to right: the (sparse) input field, the MeteoSwiss ground truth, the model prediction, and the residual bias.

\begin{figure}[H]
    \centering
    \includegraphics[width=\textwidth]{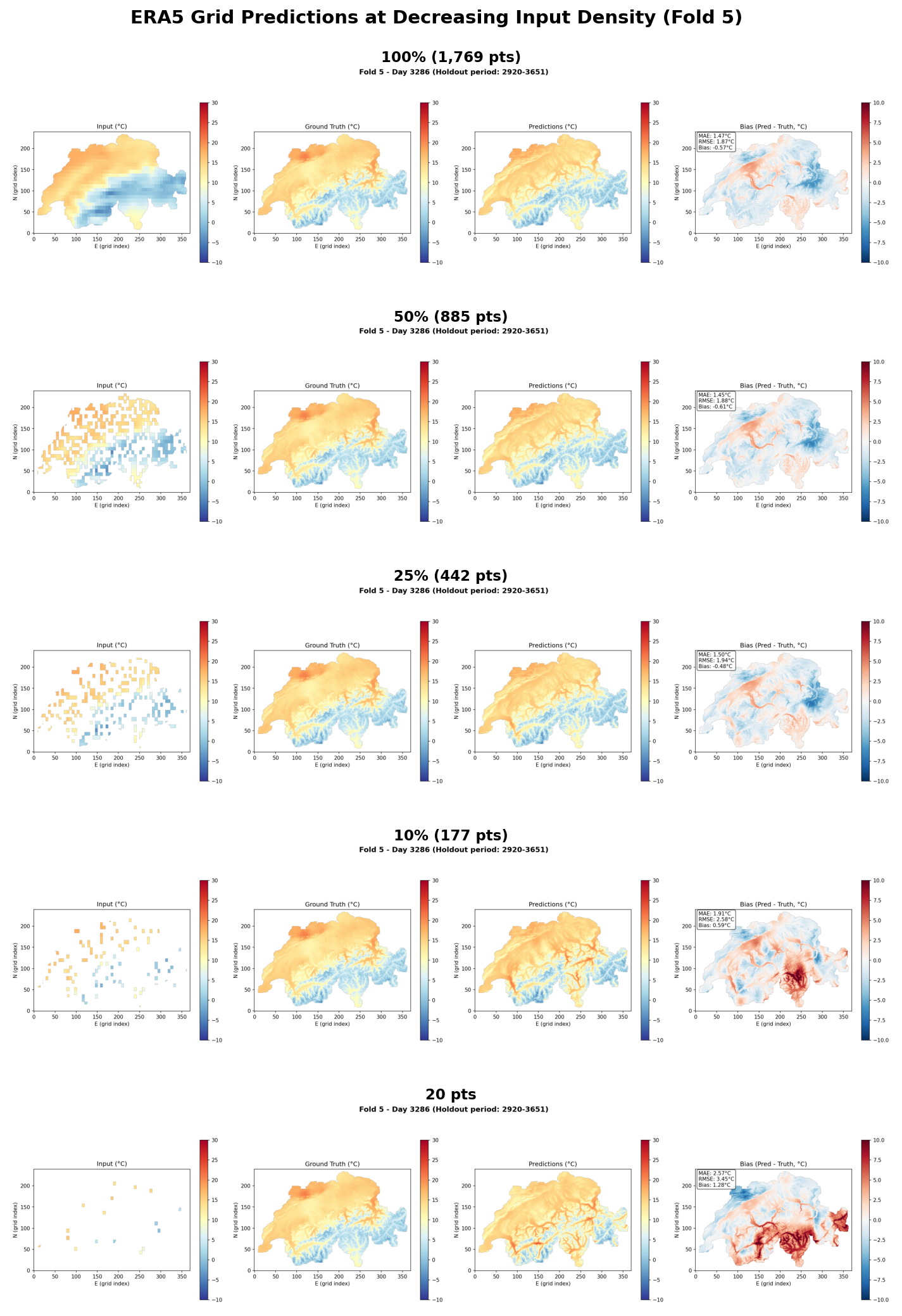}
    \caption{ERA5 on-grid predictions at decreasing input density: 100\% (1\,769~pts), 50\% (885~pts), 25\% (442~pts), 10\% (177~pts), and 20~pts. The input field becomes visibly sparser while the model attempts to maintain spatial coherence; at 20~points the prediction degrades substantially.}
    \label{fig:sparse_era5_predictions}
\end{figure}

\begin{figure}[H]
    \centering
    \includegraphics[width=\textwidth]{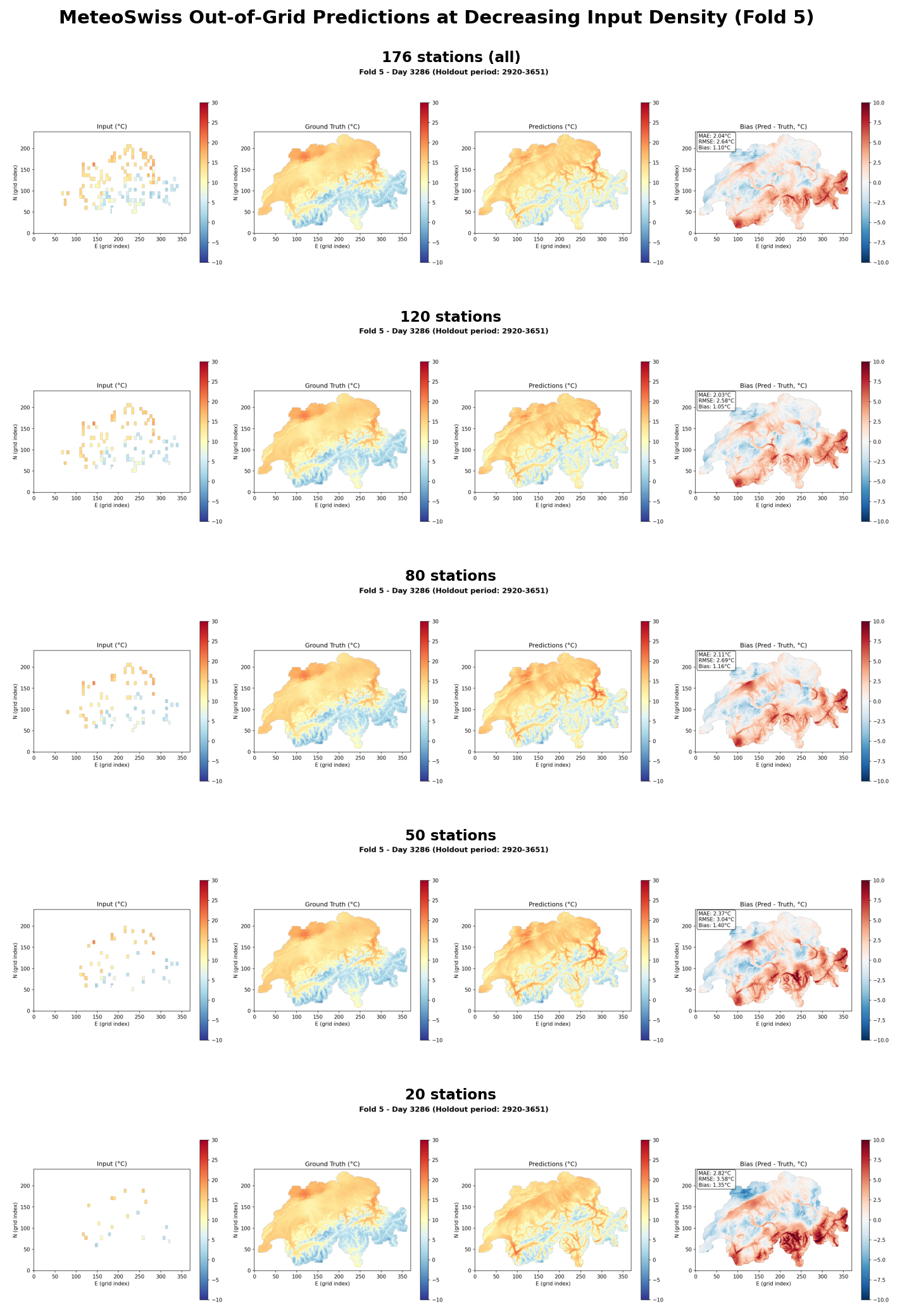}
    \caption{MeteoSwiss off-grid predictions at decreasing station count: 176 (all available), 120, 80, 50, and 20~stations. The sparse station observations are snapped to the nearest ERA5 grid cell; large areas of the input field are unobserved (NaN). A warm bias is visible across all configurations.}
    \label{fig:sparse_meteoswiss_predictions}
\end{figure}

\section{Glossary of Terms}

\begin{description}

    \item[Context set] The collection of input observations (locations and their associated values) provided to a Neural Process at inference time. In this project, the context set consists of ERA5-Land grid points with their temperature values and auxiliary features.

    \item[Density channel] An auxiliary channel in the ConvCNP's set convolution output that records the observation density at each grid cell. It informs the CNN about where context data is present and where it is absent, allowing the model to increase uncertainty in data-sparse regions.

    \item[Downscaling (statistical)] The process of deriving high-resolution local predictions from coarse-resolution gridded data by learning statistical relationships between large-scale fields and local observations, as opposed to dynamical downscaling which runs a high-resolution physics-based model.

    \item[Gaussian Process (GP)] A stochastic process in which any finite collection of random variables follows a multivariate Gaussian distribution, fully specified by a mean function and a covariance (kernel) function. GPs provide analytically tractable uncertainty estimates but scale as $\mathcal{O}(N^3)$ with the number of observations, which limits their applicability to large spatial datasets.

    \item[Lapse rate] The rate at which air temperature decreases with increasing altitude. The standard environmental lapse rate is approximately 6.5\,\textdegree C per 1000\,m, but actual lapse rates vary with season, weather conditions, and local topography (e.g.\ temperature inversions in winter valleys).

    \item[Meta-learning] A learning paradigm in which a model is trained across a distribution of tasks so that it learns not a single fixed function, but a \textit{family} of functions---captured as a predictive distribution conditioned on context. Each task corresponds to a different context--target split (e.g.\ a different day's weather). At inference time, the model conditions on a novel context set and produces a full predictive distribution over target values, without requiring retraining. Exposing the uncertainty allows the model training to minimize calibration error. See \textit{Uncertainty calibration} below.

    \item[Neural Process (NP)] A family of neural network models that define distributions over functions. Given a context set of observed input--output pairs, an NP predicts the output distribution at arbitrary target locations, combining the flexibility of neural networks with the uncertainty-aware predictions of Gaussian Processes.

    \item[Convolutional Conditional Neural Process (ConvCNP)] A member of the Neural Process family that uses a set convolution to map irregularly spaced context observations onto a regular grid, followed by a convolutional neural network to produce translation-equivariant representations. It outputs predictive distributions at arbitrary target locations.

    \item[Reanalysis] A systematic approach to producing consistent, gridded datasets of atmospheric variables by combining historical observations with a numerical weather model through data assimilation. ERA5-Land is the high-resolution land component of the ERA5 reanalysis produced by ECMWF.

    \item[Set convolution] An operation that maps a set of irregularly placed observations onto a regular grid by convolving each observation with a kernel function centered at its location. This is the mechanism by which ConvCNPs handle arbitrary input configurations.

    \item[Spatial interpolation] The estimation of a variable's value at unobserved locations based on observed values at nearby locations. Methods range from simple nearest-neighbor or bilinear schemes to geostatistical approaches such as kriging and machine learning models.

    \item[Target set] The collection of locations at which predictions are requested. In this project, target locations correspond to MeteoSwiss ground-truth weather stations.

    \item[Topographic Position Index (TPI)] A measure of the relative elevation of a point compared to its surrounding terrain. Positive values indicate ridges or hilltops; negative values indicate valleys. TPI helps characterize the local topographic context that influences temperature.

    \item[Translation equivariance] A property whereby shifting the input spatially produces a correspondingly shifted output. In ConvCNPs, this is enforced by the convolutional architecture and ensures that the model treats all spatial locations consistently.

    \item[Uncertainty calibration] The degree to which a model's predicted confidence intervals match the observed frequency of outcomes. A well-calibrated model that predicts 90\% confidence intervals should contain the true value approximately 90\% of the time.

\end{description}

\end{document}